\documentclass[letterpaper,twocolumn,10pt]{article}
\usepackage{usenix}

\usepackage{amsmath,amssymb,amsfonts}
\usepackage{array}
\usepackage{booktabs}
\usepackage{graphicx}
\usepackage{multirow}
\usepackage{placeins}
\usepackage{dblfloatfix}
\usepackage{subcaption}
\usepackage[most]{tcolorbox}
\usepackage{breqn}

\graphicspath{{./}{separated_picture/}}

\newcommand{\StateAbbr}[1]{\ensuremath{\mathit{#1}}}
\newcommand{\StateTerm}[1]{#1}

\newtcolorbox{stateexamplebox}[1][]{%
  enhanced,
  breakable,
  colback=gray!3,
  colframe=black!25,
  boxrule=0.45pt,
  colbacktitle=gray!62,
  coltitle=white,
  fonttitle=\bfseries,
  title=#1,
  arc=2pt,
  left=4pt,
  right=4pt,
  top=3pt,
  bottom=3pt,
  before skip=3pt,
  after skip=3pt
}

\hypersetup{
  hidelinks,
  pdfauthor={},
  pdfcreator={},
  pdfproducer={},
  pdftitle={Uncovering Why LLMs Fail: From Internal Mechanisms to Adversarial Reasoning Trajectories}
}

\begin{document}
%-------------------------------------------------------------------------------

\date{}

\title{\Large\bf
Uncovering Why LLMs Fail: From Internal Mechanisms to Adversarial Reasoning Trajectories
}

% Keep an anonymous author block in the review version. Do not paste names,
% affiliations, acknowledgments, grants, or institution-identifying text here.
% Camera-ready / non-anonymous author block.
% For the anonymous review version, keep {\rm Anonymous Submission} instead.
\author{
Zirui Ding\textsuperscript{1,2},
Xinxin Fan\textsuperscript{2,3},
Hao Yuan\textsuperscript{1,4},\\
Shaoye Luo\textsuperscript{2,3},
Men Niu\textsuperscript{1,2},
Quanliang Jing\textsuperscript{1},
Yunfeng Lu\textsuperscript{5}\\[4pt]
\textsuperscript{1}Institute of Computing Technology, Chinese Academy of Sciences\\
\textsuperscript{2}University of Chinese Academy of Sciences\\
\textsuperscript{3}State Key Laboratory of AI Safety, Institute of Computing Technology, Chinese Academy of Sciences\\
\textsuperscript{4}ShanghaiTech University\\
\textsuperscript{5}Beihang University
}

\maketitle

%-------------------------------------------------------------------------------
\begin{abstract}
Large language models (LLMs) are usually evaluated by final-output correctness or safety, yet an endpoint label does not reveal how a failure develops throughout the reasoning process. We therefore ask whether ordinary reasoning failures and adversarially induced failures can be characterized using the same internal coordinates. Towards this end, we in this paper study a white-box diagnostic framework \textbf{CauTrace} to disclose the root-cause of why LLM fails from three fundamental inspections: Functional States, Propagation of Representation, Temporal-State Coordination and Reasoning Trajectory. First, seven functional states are creatively defined to represent the \textit{support} and \textit{constraint} processing in LLMs. Second, the propagation of representation change measures how the changes among the seven states persist across subsequent Transformer blocks. Third, temporal-state coordination and trajectory topology characterize when the \textit{commitment} and \textit{verification} occur relative to the \textit{support}, and how the dominant state sequence evolves over the full inference process. Finally, we evaluate CauTrace on both clean reasoning failures and paired adversarial perturbations. Across Qwen and LLaMA-family models, the failed traces show weaker Attention--MLP agreement on \textit{support}-state updates and less \textit{verification} activity relative to \textit{commitment}, while the exact representation-level signatures and timing shifts vary by architecture. %Samples with the same output label can follow different trajectory topologies, and paired adversarial flips show broader changes in the same process coordinates. 
Held-out repair and steering tests demonstrate consistent operational signals that were in line with our claims for certain models and specific settings, with aligned steering producing the predicted directional changes in process measures. 
Overall, CauTrace provides a unified diagnostic framework for analyzing failure dynamics through changes in \textit{support}, \textit{commitment}, and \textit{verification}, which would shed light on the in-depth research on LLM security. %, without claiming a unique causal circuit for each failur.
\end{abstract}

%-------------------------------------------------------------------------------
\section{Introduction}
%-------------------------------------------------------------------------------
Hallucinations and harmful generations can make large language models (LLMs) appear certain while producing unsupported facts, fabricated evidence, or invalid reasoning, as documented by recent hallucination and factuality surveys \cite{huang2025survey,wang2024factuality}. In ordinary use, such outputs can propagate misinformation and mislead downstream decisions. In security-sensitive settings, HarmBench \cite{mazeika2024harmbench} and JailbreakBench \cite{chao2024jailbreakbench} standardize the measurement of whether manipulated prompts turn safety-aligned models from refusal to harmful compliance. The cost is therefore not limited to a wrong answer: a fluent but ungrounded explanation can also make the failure difficult for users or automated detectors to identify \cite{li2023halueval,manakul2023selfcheckgpt}.

Research on trustworthy and safe LLMs has identified several non-exclusive sources of these failures. TruthfulQA shows that models can reproduce popular misconceptions learned from their training distribution \cite{lin2022truthfulqa}. Lost-in-the-middle analyses \cite{liu2024lostmiddle} and distraction tests \cite{shi2023large} show that task-relevant evidence can be overlooked when it is embedded in long or irrelevant context. Calibration studies examine whether models can estimate the correctness of their own answers \cite{kadavath2022language,yin2023large}, whereas SelfCheckGPT detects unsupported passages through disagreement among sampled responses \cite{manakul2023selfcheckgpt}. Reasoning traces introduce another gap: biasing features can change an answer without being acknowledged in the generated chain of thought \cite{turpin2023language}, and intervention-based studies find that textual rationales do not always mediate the final prediction \cite{lanham2023measuring,paul2024making}. In safety-aligned models, adversarial prompts can redirect intermediate hidden-state processing \cite{zhou2024alignment}, exploit safety signals concentrated in the chat-template region \cite{leong2025template}, alter refusal-related representations \cite{arditi2024refusal}, or hijack attention from safety-relevant template tokens \cite{bentov2026attention}. These findings suggest that hallucination and unsafe compliance can involve interacting failures of evidence use, reasoning coordination, and internal control rather than one isolated defect.

Nevertheless, LLM trustworthiness and safety are still commonly evaluated from endpoint outputs, and an endpoint label alone does not reveal how the result was produced through successive Transformer layers and generation steps \cite{wang2024factuality,zhou2024alignment}. Prior studies show that the same endpoint failure label can arise from heterogeneous failure patterns: task-relevant premises may be overlooked when evidence is embedded in long or distracting contexts \cite{liu2024lostmiddle,shi2023large}; parametric falsehoods or unsupported associations may dominate a response \cite{lin2022truthfulqa}; multi-hop reasoning may omit a required inference or fail to integrate the final supporting fact \cite{gupta2025novelhopqa,paul2024making}; and explicit task constraints may be violated \cite{jiang2024followbench}. Moreover, answer \textit{commitment} may be driven by shortcut features rather than by the stated \textit{support}, with the reported reasoning merely rationalizing that choice \cite{turpin2023language}; a later self-check may then fail to repair the answer or may even degrade it \cite{huang2024selfcorrect}. The same ambiguity arises in safety evaluation: an unsafe completion establishes that a safeguard failed, but not which part of the internal generation process was reorganized by the adversarial prompt \cite{chao2024jailbreakbench,zhou2024alignment}. Consequently, endpoint evaluation alone leaves a fundamental diagnostic question unanswered: how can failures be traced to changes within the model's internal generation pipeline?

Existing research therefore provides four important but partial analytic views. Endpoint benchmarks measure factuality through learned misconceptions in TruthfulQA \cite{lin2022truthfulqa}, generated and human-annotated hallucinations in HaluEval \cite{li2023halueval}, or standardized harmful-behavior evaluation in HarmBench \cite{mazeika2024harmbench} and JailbreakBench \cite{chao2024jailbreakbench}. Calibration and consistency methods estimate whether an answer is unreliable \cite{kadavath2022language,manakul2023selfcheckgpt}; chain-of-thought studies test whether textual rationales faithfully mediate predictions \cite{turpin2023language,paul2024making}; and mechanistic methods decode hidden representations \cite{belrose2023eliciting,ghandeharioun2024patchscopes} or localize safety-relevant directions \cite{arditi2024refusal}. These views are rarely connected by a common process-level unit. Endpoint scores can merge distinct failure routes, textual rationales need not expose the operative computation, and a localized representation does not by itself explain when support, commitment, and verification become miscoordinated over an entire generation. A unified analysis therefore needs to connect \emph{where} a change persists, \emph{when} functional roles switch or lose coordination, and \emph{which} whole-trace route produces the endpoint.

In this work, our goal is therefore not to propose another endpoint detector or a stronger attack, but to ask a nontrivial question: \textbf{Can non-adversarial reasoning failure and adversarially-induced reasoning failure be described with a common set of reasons?} Surrounding this question, we treat a well-formed generation as a reasoning trajectory in which task-relevant \textit{support} and \textit{constraint} are established before answer commitment becomes dominant, and later \textit{verification} remains able to confirm or revise the candidate answers. Failures can therefore appear as weak \textit{support} continuity, shifted \textit{commitment}/\textit{verification} timing, frequent state switching, or low verification activity relative to \textit{commitment}, which offers an operational and quantitative criterion for analyzing reasoning failures.

%Fig.~\ref{fig:overview} summarizes the logic of CauTrace. 
Generally, the final LLM-output answers whether the response is incorrect, unsupported, or unsafe, but cannot locate which part of the internal process changed; the same output label can also hide different state-transition routes. Our proposed CauTrace first maps each generation to seven functional-state coordinates describing support/constraint processing, answer commitment, and verification. 
It then analyzes three complementary properties: whether representation changes persist across subsequent Transformer blocks, whether the timing of \textit{support}, \textit{commitment}, and \textit{verification} remains coordinated, and which trajectory topology emerges over the full generation trace. %Paired clean/adversarial prompts keep the task objective fixed and provide a controlled stress test of whether the same process coordinates change when behavior flips. 
In a nutshell, three main contributions are involved:
\begin{itemize}
    \item \textbf{Reasoning-Process Formulation.} We creatively define seven radical functional-state coordinates to abstract the entire reasoning process, by which the support/constraint processing, answer commitment, and verification can be formulated appropriately.
    \item \textbf{Three-Level Failure-Diagnostic Analytics.} We construct the propagation of representation mechanism across Transformer blocks, temporal coordination among functional states, and whole-trace trajectory topology, by which the fine-grained representation changes can be captured microscopically.
    %\item \textbf{Empirical separation and adversarial stress testing.} Across Qwen and LLaMA-family models, we show that failed traces differ from successful traces in support-state update agreement and relative verification activity, that samples with the same output label can follow different trajectory routes, and that paired adversarial flips are accompanied by larger internal changes.
    %\item \textbf{Explicit evidence boundary.} We use cross-family checks, aggregation sensitivity, and controlled activation patching to distinguish reproducible diagnostic associations from architecture-specific representation-level signals and from stronger causal-localization claims that the current evidence does not establish.
    \item \textbf{Empirical Verification.} Multi-facet experiment-oriented evidences across the popular Qwen and LLaMA-family models demonstrate that: i) the failed traces differ from successful traces in \textit{support}-state update agreement and relative \textit{verification} activity; ii) that even the same output label can follow different reasoning trajectory chains; and iii) that harmful-benign-paired adversarial flips are accompanied by larger internal-state change.
    \end{itemize}

%-------------------------------------------------------------------------------
\section{Related Work}
%-------------------------------------------------------------------------------
\subsection{Reliability and Reasoning Analytics}
Hallucination analytics first requires an operational definition of failure. Surveys distinguish factuality, faithfulness to a source, evidence support, and task correctness as related but non-equivalent targets \cite{ji2023survey,wang2024factuality,huang2025survey,zhang2025siren}. At the benchmark level, TruthfulQA tests whether models reproduce common misconceptions learned from human text \cite{lin2022truthfulqa}, whereas HaluEval combines generated and human-annotated samples to evaluate hallucination recognition across general queries and task-specific settings \cite{li2023halueval}. FActScore moves from response-level labels to claim-level analysis by decomposing long-form generations into atomic facts and checking each fact against a knowledge source \cite{min2023factscore}. Related task analytics isolate other endpoint failure dimensions: FollowBench separates fine-grained instruction constraints \cite{jiang2024followbench}, and NovelHopQA varies reasoning depth and context length to identify missed final-hop integration and long-range drift \cite{gupta2025novelhopqa}. These benchmarks improve diagnostic granularity at the output level, but the same failed label or unsupported claim can still be produced through different internal routes.

A complementary line examines whether the model itself provides signals of unreliability. Kadavath et al. \cite{kadavath2022language} derive probability-based estimates of answer correctness and knowledge confidence from model outputs \cite{kadavath2022language}; Yin et al. \cite{yin2023large} evaluate self-knowledge across answerable and unanswerable questions \cite{yin2023large}; and SelfCheckGPT \cite{manakul2023selfcheckgpt} uses disagreement among stochastic generations as a black-box signal of unsupported content \cite{manakul2023selfcheckgpt}. Verification is not automatically corrective, however, intrinsic self-correction without external feedback can fail to improve reasoning and can even reduce accuracy \cite{huang2024selfcorrect}. These methods provide uncertainty or consistency scores, but they do not identify which stage of evidence use, answer formation, or checking produced the score.

Reasoning analytics uses generated intermediate text as a more detailed behavioral trace. Scratchpads externalize intermediate computation \cite{nye2021show}; chain-of-thought and zero-shot reasoning prompts elicit multi-step rationales \cite{wei2022chain,kojima2022large}; and self-consistency aggregates answers across multiple sampled chains \cite{wang2023selfconsistency}. Faithfulness research separates the plausibility of a rationale from whether it actually mediates the prediction \cite{jacovi2020towards}. Biasing features can systematically change answers while remaining absent from the stated rationale \cite{turpin2023language}; rationale perturbation and truncation tests reveal task- and model-dependent faithfulness \cite{lanham2023measuring}; and causal mediation analysis finds that models do not always rely on their own intermediate steps when producing the final answer \cite{paul2024making}. Consequently, textual reasoning traces provide valuable behavioral evidence but cannot be treated as direct records of internal computation.

\subsection{Adversarial-Prompt and Safety Analytics}
Security analytics first characterizes attacks and their behavioral endpoints. HarmBench standardizes harmful behaviors, red-teaming methods, and refusal classifiers within one evaluation protocol \cite{mazeika2024harmbench}; JailbreakBench defines a reproducible threat model and artifact repository for comparing jailbreak attacks and defenses \cite{chao2024jailbreakbench}; and StrongREJECT grades both harmfulness and response informativeness so that empty or non-responsive outputs are not counted as successful jailbreaks \cite{souly2024strongreject}. Many-shot jailbreaking demonstrates that long contexts containing many demonstrations can progressively erode refusal behavior \cite{anil2024manyshot}. Corpus-based studies characterize in-the-wild jailbreak prompts \cite{shen2024doanything} and analyze transferable semantic attack patterns \cite{yu2024dontlisten}, whereas AutoDAN automates the search for stealthy jailbreak instructions \cite{liu2024autodan}. Prompt-injection work separately formalizes attacker objectives and evaluation settings \cite{liu2024promptinjection}, including optimization attacks against LLM-as-a-judge systems \cite{shi2024optimization}. This line establishes whether an attack succeeds and under which threat model, but its principal analytic unit remains the prompt--response outcome.

A second line analyzes the internal organization of safety behavior. Representation Engineering extracts population-level activation directions associated with behavioral concepts \cite{zou2023repeng}. Arditi et al. identify a direction that mediates refusal and validate it through activation addition and ablation \cite{arditi2024refusal}; subsequent studies describe refusal using concept cones \cite{wollschlager2025geometry} or multiple approximately orthogonal safety dimensions \cite{pan2025hidden}. Intermediate-state analysis compares aligned and jailbroken generations across layers \cite{zhou2024alignment}. More localized studies show that safety decisions can be anchored in the chat-template region \cite{leong2025template} and that universal suffixes redirect attention away from safety-relevant template tokens \cite{bentov2026attention}. Activated-concept analysis has also been used for jailbreak defense \cite{zhang2025jbshield}, while PromptLocate localizes prompt-injection influence within the model pipeline \cite{jia2026promptlocate}. These methods reveal safety-relevant representations or information-flow vulnerabilities, but most are organized around a particular refusal mechanism, attack family, or localized component rather than a task-general generation process.

\subsection{Mechanistic and Representation Analytics}
White-box analytics supplies the primitives for reading and intervening on model internals. Contrast-consistent learning extracts latent truth-related structure without labeled answers \cite{burns2023discovering}, while the Tuned Lens decodes layer-wise predictive distributions and exposes how token predictions evolve through the residual stream \cite{belrose2023eliciting}. Patchscopes generalizes projection and intervention methods by transplanting hidden representations into a target prompt so that the model can express their information in natural language \cite{ghandeharioun2024patchscopes}. TransformerLens \cite{nanda2022transformerlens} provides reusable instrumentation for caching, decomposing, and patching activations \cite{nanda2022transformerlens}. Together, these methods determine what information is decodable at a layer or how an intervention changes a selected behavior.

Mechanistic studies then seek smaller computational units. Feed-forward layers can operate as key--value memories that promote factual concepts \cite{geva2021transformer}; induction-head analyses identify attention circuits that implement repeated-pattern completion \cite{olsson2022context}; and causal tracing with ROME localizes and edits factual associations \cite{meng2022locating}. Automated Circuit Discovery systematizes activation-patching workflows by searching for the edges that form a behavior-specific circuit \cite{conmy2023acdc}. Such methods provide component- or circuit-level explanations, but their units are usually selected for one behavior and do not by themselves describe how functional roles coordinate over an entire generation trace.

\textbf{Summary \& Motivation.} Existing analytics offer increasingly precise endpoint scores, attack measurements, hidden-representation decoders, and localized safety mechanisms, but rarely connect these resolutions into one generation-level account of ordinary and adversarial failures. CauTrace addresses this gap by organizing a white-box evidence chain around seven functional-state/process coordinates and jointly analyzing Propagation of Representation, Temporal-State Coordination, and Reasoning Trajectory, rather than proposing another endpoint benchmark, probe, or refusal direction.

\begin{figure*}[t]
    \centering   \includegraphics[width=0.95\textwidth,height=0.37\textheight]{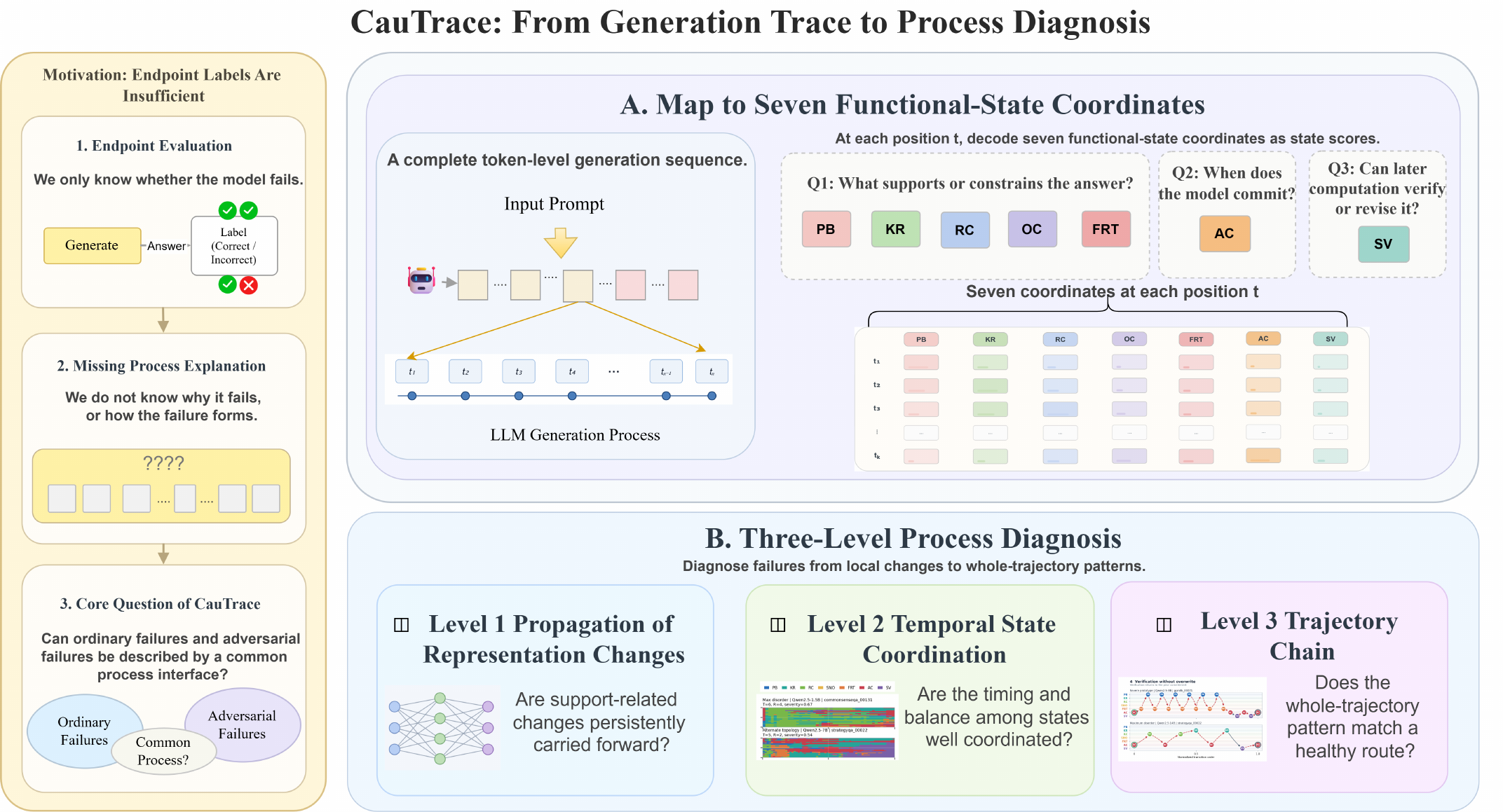}
    \caption{CauTrace overview. %Final-output labels indicate whether a generation fails but not which internal process changed. 
    CauTrace maps clean/adversarial traces to seven functional-state coordinates and analyzes (1) Propagation of Representation across Transformer blocks; (2) Temporal-state Coordination among support, commitment, and verification; and (3) Reasoning Trajectory over the full generation trace.}
    \label{fig:overview}
\end{figure*}

%-------------------------------------------------------------------------------
\section{Methodology}
\label{sec:theory}
%-------------------------------------------------------------------------------
Fig. \ref{fig:overview} sketches the overview framework, the propose CauTrace treats each generation as one hidden-state trajectory and uses the seven functional-state coordinates as a reasoning chain to maps the generated trace. To unveil the inherent reasoning process, we inspect three-level linked analytics: \textbf{Propagation of Representation.} It examines whether a local change is retained or canceled/absorbed alongside Transformer's Attention and MLP updates; ii) \textbf{Temporal-State Coordination.} It is used to reconstruct how the same coordinates switch and align dynamically over the generation time, including whether \textit{support} and \textit{constraint} precede answer \textit{commitment} and whether the later \textit{verification} can revise it back; and iii) \textbf{Reasoning Trajectory.} It integrates the local and temporal measurements into a whole-trace route for further observation and judgement on the output trajectory.Together, these three levels provide a unified view of the reasoning process across different analytical resolutions.

The above three-level analytics form a measurement hierarchy rather than three independent feature inspections. Representation analytics aims to identify \emph{where} a deviation remains available to later computation; temporal-state analysis identifies \emph{when} state-links, commitment, and verification become miscoordinated; and trajectory chains specify \emph{which} reasoning chain exactly lead to the final results. Therefore, in view of these three-level analytics, they are hierarchical and mutually complementary to calibrate the reasoning trace.

\subsection{Functional-State Coordinates}
\label{subsec:functional-states}
Inspired by decompositions of problem-solving into representation, knowledge access, operation, monitoring, and answer confirmation \cite{davidson2003psychology,newell1972human},  we define seven functional-state/process coordinates for LLM reasoning. They address three questions: \textbf{what supports or constrains the answer?} \textbf{when does the model commit to a candidate answer?} and \textbf{can the later computation verify or revise that commitment?} Table~\ref{tab:state_definitions} explicitly maps each state to one of the three research questions (Que. I-III): \StateAbbr{PB}, \StateAbbr{KR}, \StateAbbr{RC}, \StateAbbr{OC}, and \StateAbbr{FRT} represent support or constraint formation; \StateAbbr{AC} describes answer selection; and \StateAbbr{SV} indicates post-commitment verification and revision.

\textbf{Why These Seven States.} Three considerations motivate this seven-state decomposition: i) \textbf{Coverage of Roles of LLM Reasoning.}
The seven-state decomposition covers recurring functional roles in reasoning, including binding, evidence access and composition, operational computation, rule tracking, commitment, and checking.; ii) \textbf{Operational Separability.} Our defined seven states enables to adequately decouple the entangled roles under task-grounded weak supervision; and iii) \textbf{Failure-Type Diagnostic Correction.} The seven states can validly distinguish the categories of LLM failures, also verified by the subsequent experiments. We here do not claim that these seven states are a unique or minimal cognitive decomposition. The first five states can overlap and need not occur in a fixed order. They provide a general task decomposition at a suitable granularity that can be used to deeply inspect the hidden-state reasoning process in a microscopic manner.

\begin{table*}[htbp]
\centering
\caption{Seven functional-state coordinates organized by three diagnostic questions. %Que. I concerns answer support and task constraints; Que. II concerns answer commitment; Que. III concerns post-commitment verification and revision.
}
\label{tab:state_definitions}
\begin{tabular}{>{\centering\arraybackslash}m{4.8cm}
>{\centering\arraybackslash}m{7.6cm}
>{\centering\arraybackslash}m{3.2cm}}
\hline
\textbf{Seven States} & \textbf{Functional Roles} & \textbf{Questions} \\
\hline
\textbf{Premise Binding (\StateAbbr{PB})} & Bind entities, conditions, and prompt constraints & \textbf{Que.I} \\
\hline
\textbf{Knowledge Retrieval (\StateAbbr{KR})} & Activate relevant factual or commonsense knowledge & \textbf{Que. I} \\
\hline
\textbf{Reasoning Composition (\StateAbbr{RC})} & Compose evidence across hops or intermediate claims & \textbf{Que. I} \\
\hline
\textbf{Operational Computation (\StateAbbr{OC})} & Execute numerical or symbolic operations & \textbf{Que. I} \\
\hline
\textbf{Formal Rule Tracking (\StateAbbr{FRT})} & Maintain task rules, labels, and output constraints & \textbf{Que. I} \\
\hline
\textbf{Answer Commitment (\StateAbbr{AC})} & Select and stabilize a candidate answer & \textbf{Que. II} \\
\hline
\textbf{Self-Verification (\StateAbbr{SV})} & Check, revise, or reject the committed candidate & \textbf{Que. III} \\
\bottomrule
\end{tabular}
\end{table*}

Let \StateAbbr{PB}, \StateAbbr{KR}, \StateAbbr{RC}, \StateAbbr{OC}, and \StateAbbr{FRT} respectively denote the \textit{support}/\textit{constraint} family,
\begin{equation}
\mathcal{S}^{+}=\{\StateAbbr{PB}, \StateAbbr{KR}, \StateAbbr{RC}, \StateAbbr{OC}, \StateAbbr{FRT}\}.
\end{equation}
The family captures premise binding, knowledge retrieval, reasoning composition, operational computation, and rule tracking. The defined \StateAbbr{AC} can convert these signals into a candidate answer; \StateAbbr{SV} can evaluate whether the candidate answer remains supported or revisable. Hence, the process failure is defined by abnormal coordination rather than by a missing state. A probe score $p_s(t)$ measures the evidence for an operational role in the hidden representation; it does not indicate the entry into a discrete reasoning stage. CauTrace therefore allows overlapping state scores and ostensibly "non-sequential" transitions, despite different categories of LLM failures are subject to specified reasoning trajectories.
To make these functional coordinates concrete and clear, we provide a simple reasoning example below.

\begin{stateexamplebox}[State Walkthrough Example]
\label{StateExample}
\textbf{Question.} Can a person born in 2000 legally vote in the United States in 2017?\par
\textbf{Target Output.} No\par
\textbf{State Interpretation.}
\begin{enumerate}
    \setlength{\itemsep}{1pt}
    \setlength{\parskip}{0pt}
    \setlength{\parsep}{0pt}
    \item \StateTerm{\textbf{Premise Binding}}: bind the birth year (2000), decision year (2017), and jurisdiction (United States).
    \item \StateTerm{\textbf{Knowledge Retrieval}}: recall the voting-age rule that a person must be at least 18.
    \item \StateTerm{\textbf{Reasoning Composition}}: combine the age rule with the question conditions.
    \item \StateTerm{\textbf{Operational Computation}}: compute that the person is 17 in 2017.
    \item \StateTerm{\textbf{Formal Rule Tracking}}: keep the answer in the required Yes/No decision space.
    \item \StateTerm{\textbf{Answer Commitment}}: select \emph{No} once support is sufficient.
    \item \StateTerm{\textbf{Self-Verification}}: check that the age calculation and the voting rule are consistent with \emph{No}.
\end{enumerate}
\end{stateexamplebox}

\subsection{Propagation of Representation}
\label{subsec:structural-retention}
The propagation of representation aims to analyze whether a local change in \textit{support} state signals is canceled/absorbed or persists across later Transformer updates. That is, the Attention--MLP update agreement measures whether the two sublayers move support-state in the same direction; the value-spectrum statistics measure changes in the value-projected representation of the state probes; and the alignment with the Answer Commitment (\StateAbbr{AC}) probe tests whether the dominant value-space direction becomes more commitment-related. %No single statistic is interpreted as a cause.

\textbf{Representation-Level Deviation Score.} Let $t\in\{1,\ldots,T\}$ denote a generation step, $l\in\mathcal{L}$ an analyzed Transformer layer, $\mathcal{S}$ the seven-state set and $s\in\mathcal{S}$ a specific state. Let $p^{\mathrm{in}}_{s,t,l}$, $p^{\mathrm{attn}}_{s,t,l}$, and $p^{\mathrm{out}}_{s,t,l}$ represent the state-coordinate scores at the block input, after the Attention sublayer, and after the MLP sublayer, respectively. We define
\begin{align}
\Delta^{\mathrm{attn}}_{s,t,l}
&=
p^{\mathrm{attn}}_{s,t,l}
-
p^{\mathrm{in}}_{s,t,l},
\\
\Delta^{\mathrm{mlp}}_{s,t,l}
&=
p^{\mathrm{out}}_{s,t,l}
-
p^{\mathrm{attn}}_{s,t,l},
\label{eq:sublayer-gains}
\\
A_{s,t,l}
&=
\operatorname{sign}\!\left(\Delta^{\mathrm{attn}}_{s,t,l}\right)
\operatorname{sign}\!\left(\Delta^{\mathrm{mlp}}_{s,t,l}\right),
\label{eq:module-alignment}
\end{align}
where $\Delta^{\mathrm{attn}}_{s,t,l}$ and $\Delta^{\mathrm{mlp}}_{s,t,l}$ are Attention-side and MLP-side changes in the corresponding state-coordinate score. $A_{s,t,l}\in\{-1,0,1\}$ records whether the two sublayers move the state in the same, opposite, or zero direction. We average $A_{s,t,l}$ over support states and analyzed layers to obtain $\bar A_{\mathrm{sup}}(t)$. A smaller $\bar A_{\mathrm{sup}}$ indicates lower agreement between Attention and MLP updates on support-related state scores.

For value-projected Attention, we analyze the projected representations rather than the parameter matrix. Let $\widetilde h_{j,l}$ be the normalized residual representation of token $j$ before Attention in layer $l$, and $W^V_{l,h}$ the value-projection matrix of head $h$. Thus, the head-level value vector can be defined as
\begin{equation}
v^V_{j,l,h}=\widetilde h_{j,l}W^V_{l,h}\in\mathbb{R}^{d_v},
\label{eq:value-vector}
\end{equation}
where $d_v$ is the value-head dimension. At generation step $t$, let $\mathcal{I}_t$ denote the token-layer-head units used for spectral analysis. We stack the corresponding value vectors into
\begin{equation}
V_t=\left[(v^V_{j,l,h})^\top\right]_{(j,l,h)\in\mathcal{I}_t}\in\mathbb{R}^{m\times d_v},
\label{eq:value-matrix}
\end{equation}
where $m=|\mathcal{I}_t|$. The local second-moment matrix is $\Sigma^V_t=(V_t)^\top V_t/m$. Let $\lambda^V_{1,t}\ge \lambda^V_{2,t}\ge\cdots$ denote the eigenvalues of $\Sigma^V_t$, sorted in non-increasing order. We respectively define the leading spectral concentration, normalized spectrum, and step-to-step spectral change as
%\begin{equation}
%\begin{aligned}
\begin{align}
\rho^V_t
&=\frac{\lambda^V_{1,t}}{\sum_j \lambda^V_{j,t}},\\
\pi^V_{t,k}
&=\frac{\lambda^V_{k,t}}{\sum_j \lambda^V_{j,t}},\\
D^V_{\mathrm{KL}}(t)
&=D_{\mathrm{KL}}\!\left(\pi^V_t \Vert \pi^V_{t-1}\right),
%\end{aligned}
\label{eq:value-spectrum}
\end{align}
%\end{equation}
where $\rho^V_t$ is the leading spectral concentration, $\pi^V_{t,k}$ is the normalized spectral mass, and $D^V_{\mathrm{KL}}(t)$ is the step-to-step spectral change. Let $u^V_{\mathrm{ac}}$ represent the answer-commitment probe direction in the same value space and $v^V_{1,t}$ the leading eigenvector, thus their alignment degree can be defined as

\begin{equation}
\Gamma_{\mathrm{ac},V}(t)
=
\left|
\cos
\left(u^V_{\mathrm{ac}},
v^V_{1,t}
\right)
\right|.
\label{eq:commit-value-align}
\end{equation}
The absolute cosine removes the arbitrary sign of the eigenvector. $\Gamma_{\mathrm{ac},V}(t)$ therefore measures alignment degre between the dominant value direction and the \StateAbbr{AC} probe direction. We combine the four indicators into
\begin{equation}
\begin{aligned}
R_{\mathrm{str}}(t)
=\beta_1[\delta_1-\bar A_{\mathrm{sup}}(t)]_{\hat{+}} + \beta_2[\delta_2-\rho^V_t]_{\hat{+}} \\+ \beta_3[D^V_{\mathrm{KL}}(t)-\delta_3]_{\hat{+}} + \beta_4[\Gamma_{\mathrm{ac},V}(t)-\delta_4]_{\hat{+}}
\end{aligned}
\label{eq:struct-risk}
\end{equation}
where $[\cdot]_{\hat{+}}=\max(\cdot,0)$. Thresholds are calibrated on correct reasoning traces: lower-tail thresholds are used for support-state agreement/concentration quantities, and upper-tail thresholds for change/alignment quantities. Each component is normalized by its interquartile range. We use equal weights ($\beta_k=1/4$) by default and evaluate weight sensitivity in Appendix~\ref{app:weight-sensitivity}. $R_{\mathrm{str}}$ is a composite representation-level deviation score instead of one causal factor.

\begin{center}
\fcolorbox{black}{gray!10}{\parbox{0.97\linewidth}{
\textbf{Diagnostic Claim I -- Representation Propagation.} A failure is not defined by one isolated activation difference. Rather, support-related changes can survive the Attention--MLP update sequence, remain visible in later value-projected representations, and become aligned with \StateAbbr{AC} instead of being canceled by subsequent blocks.
}}
\end{center}

\subsection{Temporal-State Coordination}
\label{subsec:transition-imbalance}
At this level analytics, the seven states are treated as "nodes" in a sophisticated stepwise-switching chain. This part is not a comparison of seven isolated score magnitudes, but to reconstruct the intrinsic correlation by which the \textit{support} and \textit{constraint} states switch into one another, the point at which the chain enters \StateAbbr{AC}, and whether the subsequent \StateAbbr{SV} activity returns to support formation or revises the committed candidate. We therefore analyze both the state-transition correlation---including repeated resets, skipped support links, and commitment/verification cycles---and their temporal-sate coordination, such as ordering, onset, persistence, and relative timing.

We operationalize the stepwise-switching reasoning chain with four complementary measurements: \emph{activation onset} identifies the first sustained appearance of a state; \emph{activity-weighted mean time} gives the center of mass of its score curve, \emph{dominant-state transitions} record the directed links and switching frequency, and \emph{relative verification activity} measures whether the \textit{verification} remains active in proportion to \textit{commitment}. Taken together, these statistics can appropriately describe the ordering, persistence, and revision opportunity of the reasoning chain.

Let $p_s(t)$ be the causal-pooled probe score of state $s$ at generation step $t$, and let $\theta_s$ denote state-specific activation threshold. The thresholds are calibrated on the clean-sample reasoning split and fixed before held-out or adversarial evaluation. The reasoning onset of $s$ is set as
\begin{equation}
\tau_s=\min\{t\mid p_s(t)\ge\theta_s,\;p_s(t+1)\ge\theta_s\}.
\label{eq:onset}
\end{equation}
The two-step condition suppresses the transient-score spikes. The dominant state sequence is $c_t=\arg\max_s p_s(t)$; and its switch count and aggregate support signal are defined as
%\begin{equation}
\begin{align}
\Omega=\sum_{i=2}^{T}\mathbf{1}[c_i\neq c_{i-1}],
\\%\qquad
S(t)=\frac{1}{|\mathcal{S}^{+}|}\sum_{s\in\mathcal{S}^{+}}p_s(t).
\label{eq:osc-support}
\end{align}
%\end{equation}

For any nonnegative signal $u=(u_1,\ldots,u_T)$, we define its center of mass as
\begin{equation}
\operatorname{CoM}(u)=\frac{\sum_{t=1}^{T}t\,u_t}{\sum_{t=1}^{T}u_t+\epsilon},
\end{equation}
where $\epsilon>0$ is a small stabilizer. Let $p_{\mathrm{com}}(t)=p_{\StateAbbr{AC}}(t)$ and $p_{\mathrm{ver}}(t)=p_{\StateAbbr{SV}}(t)$. The aggregate-support onset $\tau_{\mathrm{sup}}$ is computed by applying the same two-step rule to $S(t)$ with threshold $\theta_{\mathrm{sup}}$, calibrated like $\theta_s$. We quantify commitment/support timing, verification/commitment timing, and relative verification activity by
%\begin{equation} 
%\begin{aligned}
\begin{align}
g_{cs}&=\operatorname{CoM}(p_{\mathrm{com}})-\operatorname{CoM}(S),\\
g_{vc}&=\operatorname{CoM}(p_{\mathrm{ver}})-\operatorname{CoM}(p_{\mathrm{com}}),\\
R_{vc}&=\frac{A_{\mathrm{ver}}}{A_{\mathrm{com}}+\epsilon},
\quad
A_x=\frac{1}{T}\sum_{t=1}^{T}x_t,
%\end{aligned}
\label{eq:gaps}
\end{align}
%\end{equation}
where $g_{cs}<0$ is the early-commitment direction, while larger positive $g_{vc}$ indicates later verification relative to commitment. $R_{vc}$ is the verification-to-commitment activity ratio. Correspondingly, we can define the aggregate reasoning process-deviation score as
\begin{equation}
\begin{aligned}
R_{\mathrm{proc}}={}&\alpha_1\mathbf{1}[\tau_{\mathrm{com}}<\tau_{\mathrm{sup}}]
+\alpha_2[-g_{cs}]_+
+\alpha_3\frac{\Omega}{T}\\
&+\alpha_4[g_{vc}]_+
+\alpha_5[1-R_{vc}]_+,
\end{aligned}
\label{eq:process-deviation}
\end{equation}
where $\tau_{\mathrm{com}}$ is the onset of \StateAbbr{AC}.Continuous risk terms are normalized to [0,1] using the 5th and 95th percentiles of correct reasoning traces and then clipped. We choose percentile-based scaling rather than min-max normalization to reduce sensitivity to rare outliers while preserving the typical variation of successful reasoning processes.The default weights are uniform ($\alpha_k=1/5$). The score above combines early-commitment, support--commitment timing inversion, frequent dominant-state switching, late verification, and low verification activity relative to commitment.
%Thresholds and normalization constants are fixed within each experiment, and scores from independently calibrated benchmarks are not compared numerically across tables. Thus, the reported $R_{\mathrm{proc}}$ is bounded in $[0,1]$ and treats early commitment, support-commitment inversion, oscillation, late verification, and insufficient relative verification activity as comparable diagnostic factors rather than as learned free parameters. Section~\ref{subsec:weight-sensitivity} tests component ablations and randomized-weight perturbations.

\begin{center}
\fcolorbox{black}{gray!10}{\parbox{0.97\linewidth}{
\textbf{Diagnostic Claim II -- Temporal-State Coordination.} Reliability of LLM reasoning depends on the coordination of the state chain, not on the magnitude of any one single-state. Failed traces reorganize the handoffs among support/constraint states, \StateAbbr{AC}, and \StateAbbr{SV}; the onset shifts, temporal gaps, and switching patterns remain model- and failure type-dependent.
}}
\end{center}

\subsection{Reasoning Trajectory}
\label{subsec:trajectory-labels}
To draw the reasoning state microscopically, we summarize each generation as a reasoning trajectory-feature vector:
\begin{equation}
\phi(x_{1:T})=
\bigl(\tau_{\mathrm{sup}}, \tau_{\mathrm{com}}, \tau_{\mathrm{ver}}, \Omega, g_{cs}, g_{vc}, R_{vc}, L_{\mathrm{rule}}, \bar S, T \bigr),
\label{eq:trajectory-features}
\end{equation}
where $x_{1:T}$ is the sequence of state-coordinate scores, $\tau_{\mathrm{ver}}=\tau_{\StateAbbr{SV}}$, $L_{\mathrm{rule}}$ is the duration of weak rule/operation tracking, $\bar S=A_S$ denotes the mean support strength, and $T$ stands for the generation length. The feature groups correspond to support ($\tau_{\mathrm{sup}},\bar S$), commitment ordering ($\tau_{\mathrm{com}},g_{cs}$), verification ($\tau_{\mathrm{ver}},g_{vc},R_{vc}$), state switching ($\Omega$), constraint preservation ($L_{\mathrm{rule}}$), and continuation length ($T$).

These features characterize complementary aspects of trajectory organization, including support formation ($\tau_{\mathrm{sup}},\bar S$), commitment coordination ($\tau_{\mathrm{com}},g_{cs}$), verification behavior ($\tau_{\mathrm{ver}},g_{vc},R_{vc}$), transition stability ($\Omega$), constraint preservation ($L_{\mathrm{rule}}$), and continuation dynamics ($T$).

Based on these trajectory coordinates, a well-organized reasoning process is expected to satisfy three general organization principles. 
First, support-related states should provide sufficient evidence before answer commitment, indicating that candidate selection is grounded in accumulated support rather than premature commitment. 
Second, functional-state transitions should remain coordinated, with limited unnecessary oscillation or repeated revisits among unrelated states. 
Third, post-commitment verification should preserve the ability to confirm or revise the current candidate, enabling later computation to correct earlier deviations.

These principles define the expected organization of reasoning trajectories and provide a reference for identifying those abnormal transition patterns. 
In Section~\ref{subsec:trajectory-topology-evidence}, we empirically investigate how failed traces depart from these organization principles and characterize recurring trajectory patterns in the proposed feature space.

\begin{center}
\fcolorbox{black}{gray!10}{\parbox{0.97\linewidth}{
\textbf{Diagnostic Claim III -- Trajectory Heterogeneity.} An endpoint label does not identify a unique failure route. Trajectory Topology preserves differences in support continuity, commitment ordering, rule preservation, state recurrence, verification effectiveness, and continuation length, thereby separating multiple reasoning-process organizations within the same output category.
}}
\end{center}

%-------------------------------------------------------------------------------
\section{Experiment Evaluation}
%-------------------------------------------------------------------------------

\subsection{Configuration}
\label{subsec:diagnostic-procedure}
\textbf{Procedure.} CauTrace is evaluated as an offline diagnostic framework. For each generation, we (1) collect hidden states and module representations without changing decoding, (2) apply fixed probes to obtain probe score $\{p_s(t)\}_{s,t}$, (3) compute the representation-level, temporal-state, and trajectory diagnostics from Section~\ref{sec:theory}, and (4) compare successful and failed generation traces, matched prompt pairs, and model families under fixed development-set calibration. The resulting output is a process-level diagnostic profile rather than a new endpoint classifier.

The following experiments are conducted surrounding four research questions that correspond to the aforementioned three diagnostic claims:
\begin{itemize}
    \item \textbf{RQ1 -- Categories of LLM Failures:} How do failed and successful traces differ in propagation of representation and in the coordination of support, answer commitment, and self-verification?
    \item \textbf{RQ2 -- Specificity of Reasoning Trajectory:} Can the samples with the same output label be distinguished by their reasoning trajectory topology?
    \item \textbf{RQ3 -- Verification of Adversarial Mechanism:} When a harmful prompt-perturbation causes a behavioral flip, are associated the internal changes larger than that in non-flipped pairs and broadly consistent with the targeted or downstream functional states in theory?
    \item \textbf{RQ4 -- Generalization of Methodology:} Are the diagnostic analytics stable to aggregation choices, and transferable across different model families?
\end{itemize}

\subsection{Threat Model}
\label{subsec:threat-model}
For RQ3, we consider a prompt-channel adversary who controls the user prompt and task context. The attacker may introduce misleading premises, retrieval decoys, reordered reasoning cues, incorrect intermediate computations, rule-priority overrides, forced answer commitments, or instructions that suppress verification. An attack succeeds when a clean-successful reasoning instance becomes incorrect, unsupported, or task-inconsistent, or when a safe refusal becomes unsafe or partially unsafe compliance. The task objective and evaluation target remain unchanged, enabling pairwise comparison between each adversarial trace and its clean counterpart.

The adversary cannot modify model weights, decoding code, sampling parameters, trace collection, the evaluator, or CauTrace pipeline, and has no access to hidden states or diagnostic outputs during prompt construction. We exclude training-time poisoning, weight editing, tool or retrieval-system compromise, and deployment-side exfiltration. White-box access is reserved for the post-generation analyst, who inspects hidden states, module outputs, layers, heads, and Q/K/V/O after generation; it is not part of the attack model.

\subsection{Representation-Level: Support-Change Persists Across Updates}
\label{subsec:structural-evidence}
\begin{figure*}[!t]
    \centering
    \begin{subfigure}[t]{0.32\textwidth}
        \centering
        \includegraphics[width=\linewidth, height=0.165\textheight]{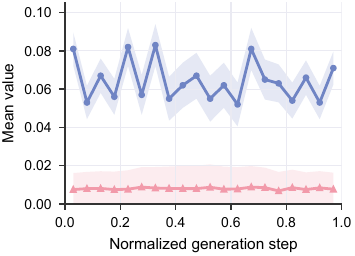}
        \caption{Value-projected spectral concentration $\rho^V_t$}
    \end{subfigure}\hfill
    \begin{subfigure}[t]{0.32\textwidth}
        \centering
        \includegraphics[width=\linewidth, height=0.165\textheight]{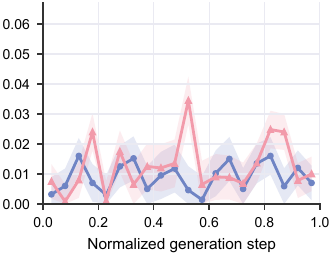}
        \caption{Value-spectrum change $D^V_{\mathrm{KL}}(t)$ }
    \end{subfigure}\hfill
    \begin{subfigure}[t]{0.31\textwidth}
        \centering
        \includegraphics[width=\linewidth, height=0.165\textheight]{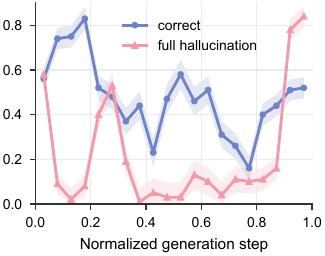}
        \caption{\mbox{Commitment alignment $\Gamma_{\mathrm{ac},V}(t)$}}
    \end{subfigure}
    \caption{Representation-level diagnostics. Panel (a) reports value-projected Attention spectral concentration $\rho^V_t$; Panel (b) reports step-to-step change in the normalized value spectrum, $D^V_{\mathrm{KL}}(t)$; Panel (c) reports alignment $\Gamma_{\mathrm{ac},V}(t)$ between the dominant value-space direction and the \StateAbbr{AC} probe direction.
    }
    \label{fig:structural_view}
\end{figure*}

To test whether local support-state changes survive later Transformer updates (\textbf{Que. I} and \textbf{Claim I}), we compare Correct and Full Hallucination traces from the nine-dataset clean reliability benchmark across Qwen2.5-1.5B/3B/7B/14B. Table~\ref{tab:structural_indicators} measures whether Attention and MLP move the support coordinates in agreeing directions, while Fig.~\ref{fig:structural_view} evaluates the value-projected Attention representation through spectral concentration, step-to-step spectral change, and alignment with the \StateAbbr{AC} probe direction. These measurements are computed from the same held-out analysis-track traces and combine probe-dependent indicators with partly probe-independent value-space statistics.

\textbf{Support-Related Representation Changes Remain Visible Across Later Transformer Updates, Despite Exact Carrier Is Not Uniform Across Scales.} As shown in Table~\ref{tab:structural_indicators}, the Attention--MLP update agreement decreases from Correct to Full Hallucination for Qwen2.5-14B ($-0.560\rightarrow-0.600$), 7B ($-0.568\rightarrow-0.625$), and 3B ($-0.500\rightarrow-0.950$); the 3B gap is the largest by $-0.450$. The 1.5B result moves in the opposite direction ($-0.700\rightarrow-0.500$), possibly because the substantially smaller model scale leads to different internal representation dynamics and update coordination patterns.This observation suggests that $\bar A_{\mathrm{sup}}$ should not be interpreted as a scale-invariant failure classifier, but rather as a mechanistic indicator of support-related propagation changes.

The value-space panels show why the conclusion is nevertheless about propagation rather than one statistic. Throughout Fig.~\ref{fig:structural_view}(a), the failed traces retain a much lower but nonzero leading value-spectrum concentration than correct traces; Fig.~\ref{fig:structural_view}(b) shows the repeated failure-associated spectral-change spikes, most visibly near the middle and final 20\% of generation; and Fig.~\ref{fig:structural_view}(c) shows failed traces becoming sharply aligned with the \StateAbbr{AC} direction only near the endpoint, whereas correct traces remain commitment-aligned across a broader portion of the trace. The three panels therefore clearly connect a changed \textit{support} representation to later \textit{commitment} organization: the deviation is not erased after its local update, but remains available and is reorganized near \textit{commitment}. Together with the cross-scale exception in Table~\ref{tab:structural_indicators}, this supports \textbf{Claim I} as a propagated diagnostic association, not a uniquely localized circuit.

\begin{table}[!t]
\centering
\caption{Representation-level indicators. Lower $\bar A_{\mathrm{sup}}$  denotes lower Attention and MLP updates to support states.}
\label{tab:structural_indicators}
%\small
\begin{tabular}{llrrrr}
\toprule
\textbf{Model} & \textbf{Label} & \textbf{$\Delta^{attn}_{sup}$} & \textbf{$\Delta^{mlp}_{sup}$} & \textbf{$\bar A_{\mathrm{sup}}$} \\
\midrule
Qwen2.5-14B  & Correct & 0.076 & -0.015 & -0.560 \\
Qwen2.5-14B  & Full Hall. & 0.125 & -0.160 & \textbf{-0.600} \\
Qwen2.5-7B   & Correct & -0.009 & 0.009 & -0.568 \\
Qwen2.5-7B   & Full Hall. & 0.019 & -0.056 & \textbf{-0.625} \\
Qwen2.5-3B   & Correct & 0.023 & 0.029 & -0.500 \\
Qwen2.5-3B   & Full Hall. & 0.041 & -0.050 & \textbf{-0.950} \\
Qwen2.5-1.5B & Correct & 0.010 & 0.039 & -0.700 \\
Qwen2.5-1.5B & Full Hall. & 0.025 & -0.060 & \textbf{-0.500} \\
\bottomrule
\end{tabular}
\end{table}

\subsection{Temporal-State Level: Coordination of Failure-Reorganizing }
\label{subsec:typed-trajectories}
\begin{figure*}[!b]
    \centering
    \begin{subfigure}[t]{0.41\textwidth}
        \centering
        \includegraphics[width=\linewidth, height=0.16\textheight]{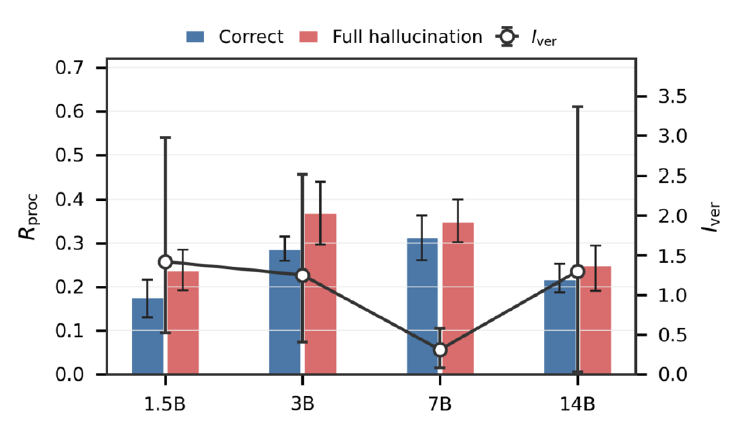}
        \caption{Composite process-deviation score across model sizes}
    \end{subfigure}\hfill
    \begin{subfigure}[t]{0.41\textwidth}
        \centering
        \includegraphics[width=\linewidth, height=0.16\textheight]{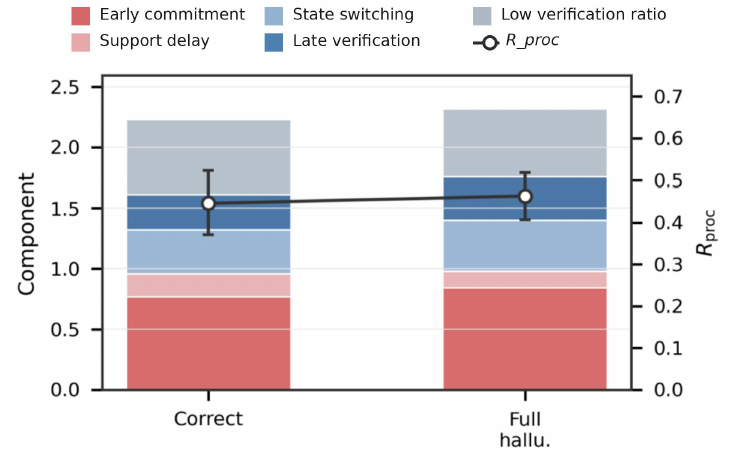}
        \caption{Process-deviation score components}
    \end{subfigure}
    \caption{Temporal-state coordination diagnostics: Panel (a) compares $R_{\mathrm{proc}}$ and the verification-onset indicator $I_{\mathrm{ver}}$; Panel (b) decomposes $R_{\mathrm{proc}}$ into early commitment onset, support delay, oscillation, late verification, and low verification coverage.
    }
    \label{fig:state_transition_evidence}
\end{figure*}

\begin{figure*}[!t]
    \centering
    \begin{subfigure}[t]{0.45\textwidth}
        \centering
        \includegraphics[width=0.91\linewidth,height=0.15\textheight]{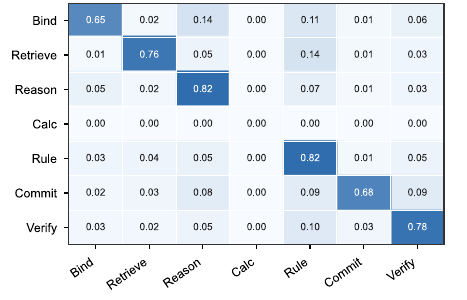}
        \caption{Correct trajectories}
    \end{subfigure}
    \begin{subfigure}[t]{0.45\textwidth}
        \centering
        \includegraphics[width=0.91\linewidth,height=0.15\textheight]{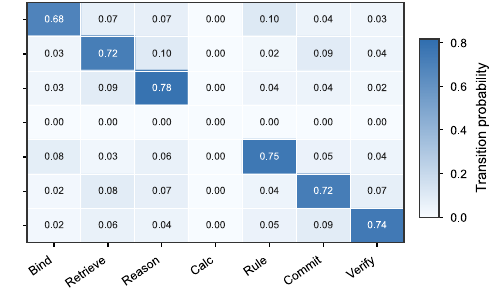}
        \caption{Failed trajectories}
    \end{subfigure}

    \begin{subfigure}[t]{0.95\textwidth}
        \centering
        \includegraphics[width=0.91\linewidth,height=0.18\textheight]{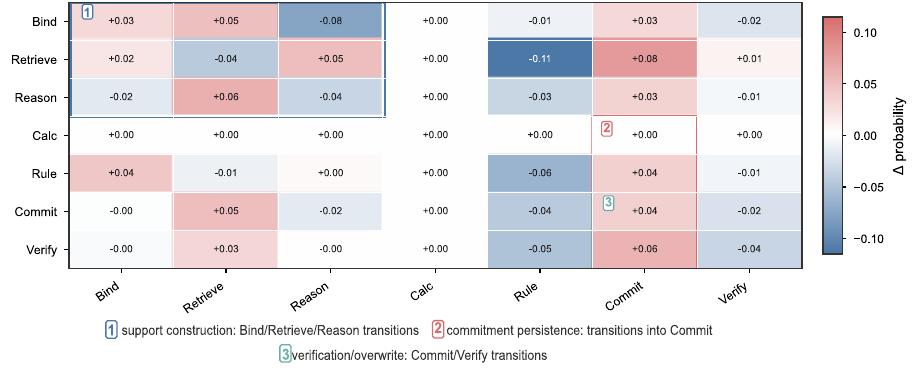}
        \caption{Failure-minus-correct shift}
    \end{subfigure}
    \caption{State-transition matrices for correct and failed trajectories: Panels (a) and (b) report transition probabilities for correct and failed generations, respectively; Panel (c) reports the failure-minus-correct difference.}
    \label{fig:main-transition-shift}
\end{figure*}

To verify \textbf{Claim II}, we compare Correct and Full Hallucination traces from the same analysis track at two resolutions. Fig.~\ref{fig:state_transition_evidence} reports the aggregate reasoning process-deviation score and its components across Qwen2.5 scales; Table~\ref{tab:process-permutation-main} tests the relative placement and activity of support, \StateAbbr{AC}, and \StateAbbr{SV}; and Fig.~\ref{fig:main-transition-shift} shows which dominant-state handoffs gain or lose probability. The combination calibrate the coordination rather than treating any one single-state score as a failure label. We compute $q$-values from two-sided unpaired permutation tests with Benjamini--Hochberg correction across all 15 comparisons.

\textbf{Failures Consistently Show Insufficient Verification After Commitment, While Their Timing Relationship Varies Across Models.}
Across models, failed traces exhibit a consistently lower verification-to-commitment ratio $R_{vc}$, indicating that verification activity becomes insufficient relative to commitment after the answer direction is formed. Pooled across models, $R_{vc}$ decreases from $4.172$ to $1.065$ ($\Delta=-3.107$, $q<0.001$) as shown in Table~\ref{tab:process-permutation-main}. This reduction is significant across all Qwen2.5 scales, with failure--correct differences of $-7.783$, $-5.292$, $-1.192$, and $-1.752$ for 1.5B, 3B, 7B, and 14B, respectively. However, the exact temporal relationship between \textit{verification} and \textit{commitment} is not identical across models: after FDR correction, the \textit{verification}--\textit{commitment} gap is not significant for 1.5B and 7B. Therefore, \textbf{Claim II} is supported by a consistent reduction in \textit{verification} coverage after \textit{commitment}, rather than by assuming a universal timing reversal across all models.

Fig.~\ref{fig:state_transition_evidence}(a) provides the aggregate view: Full Hallucination has a higher aggregate reasoning process-deviation score $R_{\mathrm{proc}}$ bar than Correct at every model size, with the largest visible separation at 3B, while the overlaid verification-onset indicator varies substantially across scales. The decomposition in Fig.~\ref{fig:state_transition_evidence}(b) further shows that the aggregate change is not a uniform rise in all five components: early-commitment and late-verification contributions increase, whereas other components change only slightly or in the opposite direction. This is precisely the coordination account in \textbf{Claim II}---different component combinations can produce failure, but inadequate checking relative to commitment remains the common relation confirmed by Table~\ref{tab:process-permutation-main}.

\begin{table*}[!t]
\centering
\caption{Timing comparisons between Correct and Full Hallucination traces. Each cell reports Correct $\rightarrow$ Full Hallucination with $\Delta=$ Full Hallucination$-$Correct. For $g_{cs}$ and $g_{vc}$, the measured quantity is a shift in relative temporal position; negative $g_{cs}$ denotes the early-commitment risk direction. Lower $R_{vc}$ indicates less verification activity relative to answer commitment.}
\label{tab:process-permutation-main}
%\scriptsize
\setlength{\tabcolsep}{3.0pt}
\renewcommand{\arraystretch}{1.13}
\begin{tabular}{@{}p{0.13\linewidth}p{0.28\linewidth}p{0.3\linewidth}p{0.24\linewidth}}
\toprule
\textbf{Model} &
\textbf{\(\boldsymbol{g_{cs}}\): support--commitment gap\newline
\emph{negative = early commitment; shift = change}} &
\textbf{\(\boldsymbol{g_{vc}}\): verification--commitment gap\newline
\emph{large shift / sign flip = disrupted checking}} &
\textbf{\(\boldsymbol{R_{vc}}\): verification/commitment ratio\newline
\emph{lower ratio = less relative verification activity}} \\
\midrule
All models &
\(14.025 \rightarrow 0.489\)\newline
\(\Delta=\textbf{-13.536}\), \(q<0.001\) &
\(-0.274 \rightarrow 16.311\)\newline
\(\Delta=\textbf{16.585}\), \(q<0.001\) &
\(4.172 \rightarrow 1.065\)\newline
\(\Delta=\textbf{-3.107}\), \(q<0.001\) \\
\addlinespace[1pt]
Qwen2.5-1.5B &
\(14.182 \rightarrow -1.432\)\newline
\(\Delta=\textbf{-15.614}\), \(q=0.034\) &
\emph{not significant after FDR correction} &
\(9.064 \rightarrow 1.281\)\newline
\(\Delta=\textbf{-7.783}\), \(q=0.011\) \\
\addlinespace[1pt]
Qwen2.5-3B &
\(24.283 \rightarrow 3.125\)\newline
\(\Delta=\textbf{-21.159}\), \(q<0.001\) &
\(4.951 \rightarrow 20.616\)\newline
\(\Delta=\textbf{15.664}\), \(q=0.011\) &
\(6.473 \rightarrow 1.181\)\newline
\(\Delta=\textbf{-5.292}\), \(q<0.001\) \\
\addlinespace[1pt]
Qwen2.5-7B &
\(3.750 \rightarrow -0.857\)\newline
\(\Delta=\textbf{-4.607}\), \(q=0.042\) &
\emph{not significant after FDR correction} &
\(1.784 \rightarrow 0.592\)\newline
\(\Delta=\textbf{-1.192}\), \(q=0.013\) \\
\addlinespace[1pt]
Qwen2.5-14B &
\(9.319 \rightarrow 0.333\)\newline
\(\Delta=\textbf{-8.986}\), \(q<0.001\) &
\(-3.940 \rightarrow 21.437\)\newline
\(\Delta=\textbf{25.376}\), \(q<0.001\) &
\(2.474 \rightarrow 0.722\)\newline
\(\Delta=\textbf{-1.752}\), \(q=0.018\) \\
\bottomrule
\end{tabular}
\end{table*}

The transition matrices identify the links underlying that aggregate shift. Relative to correct traces, Fig.~\ref{fig:main-transition-shift}(c) shows that failed traces lose probability on support-preserving handoffs, such as \StateAbbr{KR}$\rightarrow$\StateAbbr{FRT} ($-0.11$) and \StateAbbr{PB}$\rightarrow$\StateAbbr{RC} ($-0.08$), while gaining probability on \StateAbbr{KR}$\rightarrow$\StateAbbr{AC} ($+0.08$), \StateAbbr{SV}$\rightarrow$\StateAbbr{AC} ($+0.06$), and \StateAbbr{RC}$\rightarrow$\StateAbbr{KR} ($+0.06$). These selective shifts in Fig.~\ref{fig:main-transition-shift}(c) mean that failure does not merely reduce all state activity: support construction is rerouted toward commitment or reset, and verification is more likely to return to the committed candidate. This discrete link evidence explains how the lower $R_{vc}$ in Table~\ref{tab:process-permutation-main} appears as a reorganization of the state chain.

\subsection{Trajectory-Topology Level: One Output Label Contains Multiple Routes}
\label{subsec:trajectory-topology-evidence}

\begin{figure*}[!t]
    \centering
    \captionsetup[subfigure]{font=footnotesize,labelfont=bf}
    \setlength{\abovecaptionskip}{2pt}
    \begin{subfigure}[t]{1\textwidth}
        \centering
        \includegraphics[width=1\linewidth,keepaspectratio]{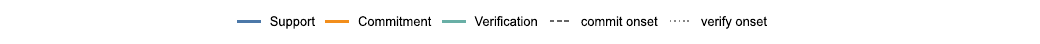}
    \end{subfigure}
    \begin{subfigure}[t]{0.29\textwidth}
        \centering
        \includegraphics[width=1\linewidth, height=0.145\textheight]{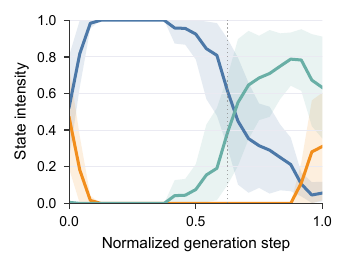}
        \caption{Normal Convergence}
    \end{subfigure}
    \hspace{0.2cm}
    \begin{subfigure}[t]{0.29\textwidth}
        \centering
        \includegraphics[width=1\linewidth, height=0.145\textheight]{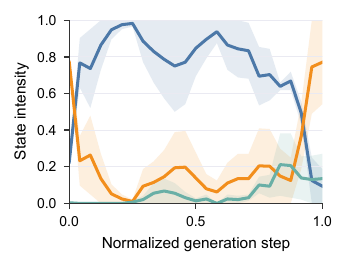}
        \caption{Early-Commitment}
    \end{subfigure}
    \hspace{0.2cm}
    \begin{subfigure}[t]{0.29\textwidth}
        \centering
        \includegraphics[width=1\linewidth, height=0.145\textheight]{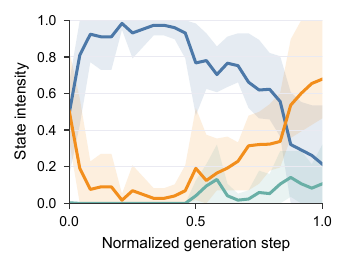}
        \caption{Fragmented Support}
    \end{subfigure}

    \begin{subfigure}[t]{0.29\textwidth}
        \centering
        \includegraphics[width=1\linewidth, height=0.145\textheight]{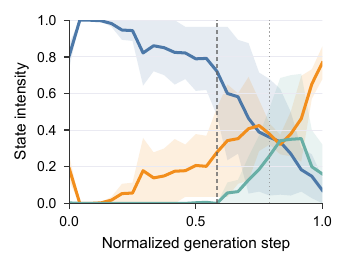}
        \caption{Rule Failure}
    \end{subfigure}
    \hspace{0.2cm}
    \begin{subfigure}[t]{0.29\textwidth}
        \centering
        \includegraphics[width=1\linewidth, height=0.145\textheight]{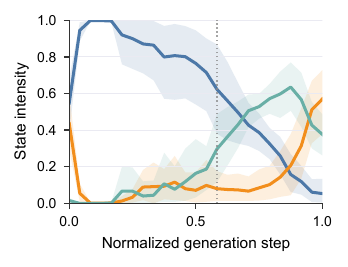}
        \caption{Verification without Overwrite}
    \end{subfigure}
    \hspace{0.2cm}
    \begin{subfigure}[t]{0.29\textwidth}
        \centering
        \includegraphics[width=1\linewidth, height=0.145\textheight]{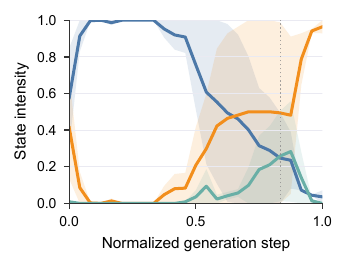}
        \caption{Verbose Unsupported Continuation}
    \end{subfigure}

    \caption{Time-domain state-score trajectories for one normal and five failure-prone trajectory types. The curves show how each trajectory pattern differs in support formation, answer commitment, and self-verification over the generation process.}
    \label{fig:app-state-curves}
\end{figure*}

\begin{figure*}[!t]
    \centering
    \includegraphics[width=0.94\textwidth,height=0.51\textheight]{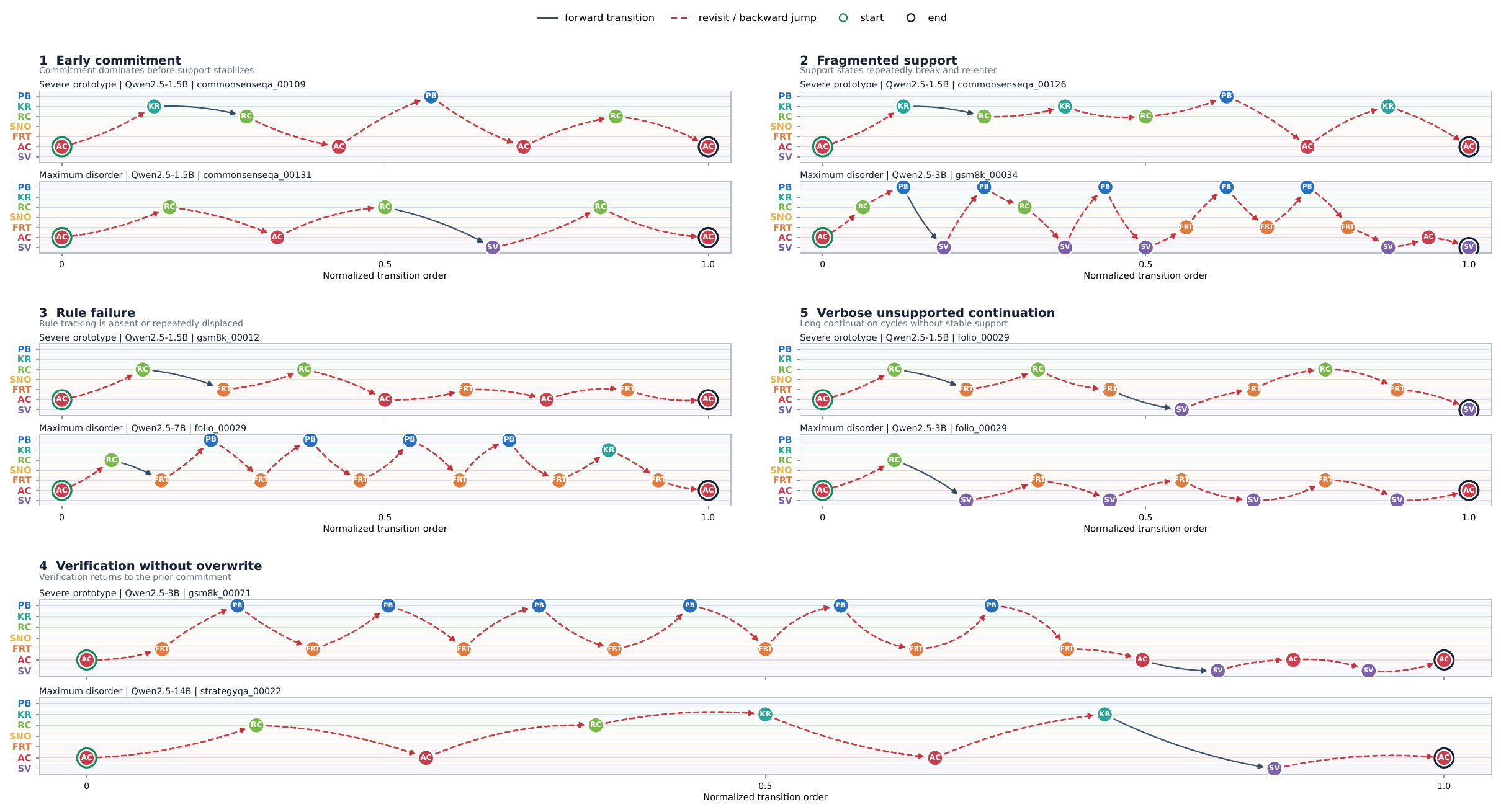}
    \caption{Representative state-transition chains for the five failure-prone trajectory types. }
    \label{fig:reasoning_state_trajectories}
\end{figure*}

To validate \textbf{Claim III}, we summarize the same held-out Full Hallucination traces with the trajectory features in Eq.~\eqref{eq:trajectory-features} and compare the five failure-prone regions defined in Table~\ref{tab:trajectory_signatures} in Appendix~\ref{app:trajectory-signatures}. Fig.~\ref{fig:trajectory_space} tests whether the regions occupy distinguishable neighborhoods in the full feature space, Fig.~\ref{fig:app-state-curves} exposes their time-domain organization, and Fig.~\ref{fig:reasoning_state_trajectories} makes the corresponding state-order motifs explicit.

\textbf{Full Hallucination Exhibits Topologically Heterogeneity: Some Routes Are Sharply Distinguishable, While Others Overlap because They Share Only Part of \textit{Support}--\textit{Commitment}--\textit{Verification} Organization.} In Fig.~\ref{fig:trajectory_space}(a), the first two projection axes explain only $15.5\%+12.7\%=28.2\%$ of trajectory-feature variance, yet Early-Commitment and Verification without Overwrite occupy visibly different neighborhoods. The row-normalized matrix in Fig.~\ref{fig:trajectory_space}(b) quantifies this asymmetry: Early-Commitment and Verification without Overwrite have diagonal proportions of $0.99$ and $0.84$, whereas Fragmented Support and Rule Failure have only $0.09$ and $0.25$ on the diagonal and are often mapped to Verification without Overwrite. The result validates \textbf{Claim III} without overstating separability: the endpoint category contains multiple routes, but the routes are not rigidly-delineated, but overlapping-existing regions. %rather than exhaustive ground-truth classes.

\begin{figure*}[!htbp]
    \centering
    \begin{subfigure}[t]{0.40\textwidth}
        \centering
        \includegraphics[width=0.79\linewidth,height=0.16\textheight]{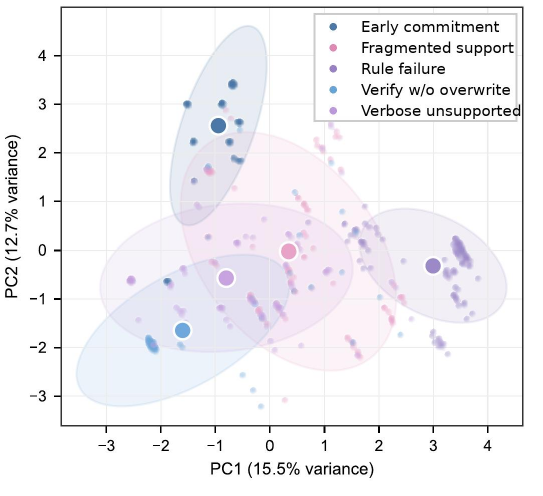}
        \caption{Trajectory-feature projection}
    \end{subfigure}
    \begin{subfigure}[t]{0.56\textwidth}
        \centering
        \includegraphics[width=0.79\linewidth,height=0.16\textheight]{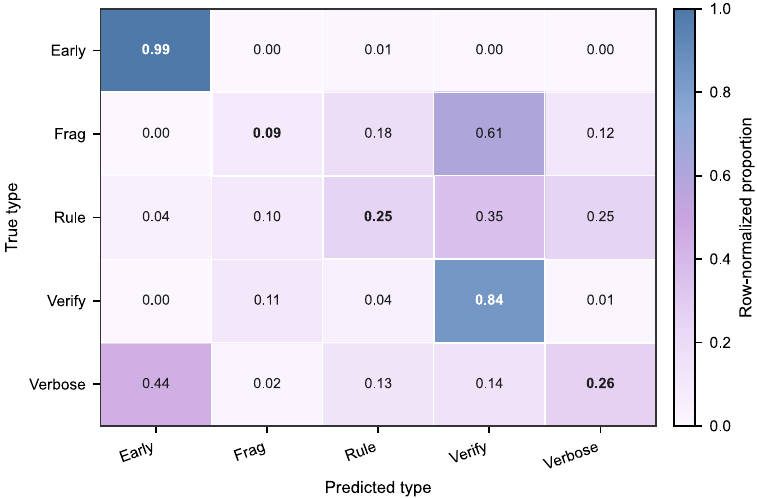}
        \caption{Type-prediction confusion matrix}
    \end{subfigure}
    \caption{Trajectory topology in the trajectory-feature space: Panel (a) visualizes trajectory-level features with approximate type neighborhoods; Panel (b) gives the row-normalized confusion matrix of a lightweight rule-derived trajectory predictor constructed directly from the diagnostic definitions in Eq.~\eqref{eq:trajectory-features}.}
    \label{fig:trajectory_space}
\end{figure*}

The time-domain panels explain what the projection preserves. Normal Convergence in Fig.~\ref{fig:app-state-curves}(a) maintains strong support through the first half of generation and then transfers activity to verification before the trace ends. Early Commitment in Fig.~\ref{fig:app-state-curves}(b) instead begins with substantial \textit{commitment} activity and returns to a strong \textit{commitment} near the endpoint; Fragmented Support in Fig.~\ref{fig:app-state-curves}(c) retains high but uneven support while \textit{verification} stays weak; and Rule Failure in Fig.~\ref{fig:app-state-curves}(d) shows \textit{support} decaying while \textit{commitment} grows without a comparably strong \textit{verification} response. The last two panels separate two superficially similar long failures: \textit{Verification} without Overwrite develops substantial late \textit{verification} while \textit{commitment} persists in Fig.~\ref{fig:app-state-curves}(e), whereas Verbose Unsupported Continuation loses support after mid-generation and ends in a renewed \textit{commitment} spike with little remaining \textit{verification} in Fig.~\ref{fig:app-state-curves}(f). These panel-specific differences interpret the temporal meaning of reasoning trajectory, i.e. the same Full Hallucination endpoint can result from premature selection, disrupted support, weak rule preservation, ineffective checking, or unsupported continuation.

Fig.~\ref{fig:reasoning_state_trajectories} translates those curve-shapes into state-order motifs. The Early-Commitment examples enter \StateAbbr{AC} before later \StateAbbr{RC}/\StateAbbr{KR} re-entry; The Fragmented Support examples repeat \StateAbbr{KR}/\StateAbbr{RC} resets; and the Verification without Overwrite examples explicitly return from \StateAbbr{SV} to \StateAbbr{AC}. The Verbose Unsupported Continuation examples are the longest and most recurrent in the figure ($T=10,R=7$ and $T=11,R=8$), consisting of repeated commitment/support re-entry without proportional verification. %Because the figure is schematic, it is used to interpret the quantitative neighborhoods and curves in Figs.~\ref{fig:trajectory_space} and~\ref{fig:app-state-curves}, not as an additional estimate of class accuracy. 
The above analytics validate \textbf{Claim III} by showing that trajectory types encode families of routes rather than merely one token-level chain per output label.

\subsection{Adversarial Test: Behavioral Flips Coincide with Larger Internal Changes}
\label{subsec:adv-process-level}

To test whether the adversarial prompt-induced failures disturb the same reasoning-process organization as clean failures (\textbf{RQ3}), we compare each adversarial trace with a matched clean trace under a fixed task objective. Table~\ref{tab:security-results} reports 500 reasoning-interference pairs and 300 AdvBench/HarmBench safety pairs per model; Table~\ref{tab:security-mechanisms} then compares state-coordinate, cascade, onset, and Q/K/V/O changes for flipped and non-flipped pairs, here a flip means that a clean correct/evidence-grounded answer becomes erroneous or unsupported, or that a clean safe refusal becomes unsafe or partially unsafe compliance. The two tables therefore separate endpoint vulnerability from the internal route associated with a successful adversarial perturbation.

\textbf{Behavioral Flips Are Accompanied by Broad Changes in Same Functional-State Coordinates.} In the reasoning-interference rows of Table~\ref{tab:security-results}, PFR ranges from $1.8\%$ to $6.2\%$ and conditional PFR from $10.0\%$ to $26.1\%$; in the safety rows, PFR is higher at $12.7\%$--$22.3\%$ and conditional PFR at $14.7\%$--$30.5\%$. These paired flips establish that the prompt intervention can change the endpoint under a matched objective. However, TAR@1 is only $15.2\%$--$16.6\%$ versus the $1/7=14.3\%$ chance level, and TAR@3 is $41.8\%$--$49.8\%$ versus $3/7=42.9\%$, as shown in Table~\ref{tab:security-results}, and Appendices~\ref{app:prompt-paired} and~\ref{app:targeted-metrics} provide the paired construction, target-state taxonomy, and metric definitions. % for Table~\ref{tab:security-results}. 
We leverage the working mechanism to test related reasoning-process changes in theory, while treating exact target-state identification as unsupported.

% \begin{figure*}[!t]
%     \centering
%     \includegraphics[width=0.995\textwidth,height=0.52\textheight]{nonlinear_state_lane_chains.pdf}
%     \caption{Representative state-transition chains for the five failure-prone trajectory types. Each type may contain multiple state-order patterns; the figure shows representative chains consistent with its operational signature and the observed failure-minus-correct transition-probability shifts. Red arrows mark failure-associated transitions, while repeated states expose support re-entry, reset, commitment persistence, or verification-to-commitment cycling. The chains are schematic examples rather than token-level reconstructions of every generation.}
%     \label{fig:reasoning_state_trajectories}
% \end{figure*}
%Fig.~\ref{fig:app-state-curves} in  shows reasoning-state-score trajectories for one normal and five failure-prone trajectory types. 

%The paired analysis complements the clean reliability benchmark. The clean benchmark identifies process failures that arise without an attacker. The paired benchmark tests whether prompt-side interventions can induce related failure families under matched objectives. We do not retain a separate evidence-chain table here because the following behavioral and mechanism tables already instantiate the chain from endpoint flips to state and white-box carriers.

%Appendices~\ref{app:prompt-paired} and~\ref{app:targeted-metrics} provide the paired construction, target-state taxonomy, and metric definitions for Table~\ref{tab:security-results}.

\begin{table*}[!t]
\centering
\caption{Paired adversarial test results. BSR denotes clean-side success, AFR is adversarial failure rate, PFR is pair flip rate, and conditional PFR measures flips among clean-successful pairs. Top-k Target Accuracy Rate (TAR) is reported for process reasoning perturbations, where each perturbation defines an explicit seven-state target.}
\label{tab:security-results}
%\scriptsize
%\setlength{\tabcolsep}{3.5pt}
\begin{tabular}{llrrrrrrr}
\toprule
\textbf{Dataset} & \textbf{Model} & \textbf{Pairs} & \textbf{BSR} & \textbf{AFR} & \textbf{PFR} & \textbf{Cond.PFR} & \textbf{TAR@1} & \textbf{TAR@3} \\
\midrule
Reasoning Interfered
 & Qwen2.5-1.5B & 500 & 23.8\% & \textbf{34.0\%} & \textbf{6.2\%} & \textbf{26.1\%} & 16.6\% & \textbf{49.8\%} \\
 & Qwen2.5-3B   & 500 & 18.0\% & 14.0\% & 1.8\% & 10.0\% & 15.2\% & 41.8\% \\
 & Qwen2.5-7B   & 500 & 26.6\% & 21.0\% & 4.8\% & 18.0\% & 16.6\% & 48.4\% \\
\midrule
AdvBench/HarmBench
 & Qwen2.5-1.5B & 300 & 73.3\% & \textbf{37.0\%} & \textbf{22.3\%} & \textbf{30.5\%} & -- & -- \\
 & Qwen2.5-3B   & 300 & 69.3\% & 33.3\% & 14.7\% & 21.2\% & -- & -- \\
 & Qwen2.5-7B   & 300 & \textbf{86.0\%} & \textbf{18.0\%} & 12.7\% & 14.7\% & -- & -- \\
\bottomrule
\end{tabular}
\end{table*}

\begin{table*}[!t]
\centering
\caption{Internal changes under paired adversarial perturbations. %Dominant shifts are adversarial-minus-clean averages over all pairs. Flip-associated signals report the strongest standardized flip-versus-non-flip contrast. Q/K/V/O signals are computed from 100 complete clean/adversarial pairs per model; $d_z$ denotes paired standardized change.
}
\label{tab:security-mechanisms}
%\scriptsize
%\setlength{\tabcolsep}{3pt}
\begin{tabular}{@{}p{0.10\textwidth}p{0.12\textwidth}p{0.22\textwidth}p{0.22\textwidth}p{0.22\textwidth}}
\toprule
\textbf{Dataset} & \textbf{Model} & \textbf{Dominant State Shift} & \textbf{Strongest Flip-Associated Signal} & \textbf{Strongest Paired QKV Signal} \\
\midrule
Reasoning Interfered
 & Qwen2.5-1.5B & Premise Binding $-0.133$
 & \StateAbbr{OC} cascade ($d=-3.180$)
 & K pairwise cosine $+0.003$ ($d_z=0.352$) \\
 & Qwen2.5-3B & Reasoning composition $+0.115$
 & \textbf{Formal Rule Tracking} onset delay ($d=2.980$)
 & \textbf{Q stable rank $-0.038$ ($d_z=-0.975$)} \\
 & Qwen2.5-7B & Reasoning Composition $-0.058$
 & \StateAbbr{OC} state ($d=-3.134$)
 & V--O norm correlation $+0.063$ ($d_z=0.315$) \\
\midrule
AdvBench\newline/HarmBench
 & Qwen2.5-1.5B & \textbf{Answer commitment $+0.057$}
 & Commitment cascade strength ($d=0.232$)
 & \textbf{O consecutive cosine $-0.078$ ($d_z=-1.289$)} \\
 & Qwen2.5-3B & Answer commitment $-0.001$
 & \textbf{Commitment peak-layer shift ($d=0.642$)}
 & Q pairwise-cosine dispersion $+0.006$ ($d_z=0.485$) \\
 & Qwen2.5-7B & Answer commitment $+0.002$
 & \textbf{Commitment peak-layer shift ($d=0.636$)}
 & O row-norm dispersion $-0.031$ ($d_z=-0.661$) \\
\bottomrule
\end{tabular}
\end{table*}

The flip-associated columns in Table~\ref{tab:security-mechanisms} provide the process-level support for the claim. Reasoning-interference flips show their strongest contrasts in \StateAbbr{OC} for Qwen2.5-1.5B ($d=-3.180$) and 7B ($d=-3.134$), and in \StateAbbr{FRT} onset for 3B ($d=2.980$), all in Table~\ref{tab:security-mechanisms}. These are support/constraint coordinates from \textbf{Claim II}, thus their disruption explains why a harmful-benign-paired prompt change can reroute the chain before the endpoint flips. Furthermore, safety flips follow a different route: the strongest state/cascade indicators are organized around \StateAbbr{AC}, including peak-layer shifts of $d=0.642$ for 3B and $d=0.636$ for 7B. This contrast supports a shared coordinate system, not a shared single route: attacks primarily disturb operation/rule support, whereas successful attacks are more closely associated with commitment organization.

The Q/K/V/O column in Table~\ref{tab:security-mechanisms} links those process-level changes back to \textbf{Claim I}. The largest paired representation effects are the O consecutive-cosine shift for 1.5B safety pairs ($d_z=-1.289$) and the Q stable-rank shift for 3B reasoning pairs ($d_z=-0.975$), as reported in Table~\ref{tab:security-mechanisms}. Thus, the successful adversarial perturbations are visible in both reasoning-process coordinates and fine-grained Transformer representations, while the changing Q/K/V/O carrier again argues against one universal circuit. Held-out checks in Tables~\ref{tab:guided-repair} and~\ref{tab:guided-steering} show the largest observed 1.5B reasoning reduction with unified repair ($76.0\%$ to $71.0\%$) and predicted 7B aligned-steering shifts in risk, rule, and verification. The adversarial evidence supports adversarial perturbation-linked, intervention-sensitive changes in the three-level analytics.

\subsection{Generalization}

To assess the generalizability of our proposed analytics framework, beyond Qwen2.5, we also apply the Correct/Full Hallucination analysis to LLaMA-3.2-1B/3B, LLaMA-3.1-8B, and Qwen1.5-MoE and Table~\ref{tab:generalization} reports failure-minus-correct differences in the representation-level indicators and in the verification activity relative to commitment.

\textbf{The Coordination Result Transfers Across Model Families More Consistently Than Any Individual Representation-Level Carrier.} All four models show lower support-state update agreement for failed traces, with $\Delta\bar A_{\mathrm{sup}}$ ranging from $-0.012$ to $-0.052$ as shown in Table~\ref{tab:generalization}. They also show lower "verification-to-commitment" activity, with $\Delta R_{vc}$ ranging from $-1.102$ to $-15.519$. These shared directions support the central relations as stated in \textbf{Claims I-II}. By contrast, $\Delta\rho_V$ and $\Delta\Gamma_{\mathrm{ac},V}$ change sign across models, and $\Delta D^V_{KL}$ is positive for the three dense LLaMA models but approximately zero for Qwen1.5-MoE. The cross-family results therefore support representation-propagating differences and weaker relative verification at the diagnostic level while indicating that the measurable carrier depends on architecture.

% \begin{table*}[!t]
% \centering
% \caption{Cross-family validation of representation-level and temporal diagnostics.} %Values are failure-minus-correct differences. %The results supporting the expected failure pattern: lower support-state update agreement, stronger value-space drift, and reduced relative verification activity.}
% \label{tab:generalization}
% %\footnotesize
% %\setlength{\tabcolsep}{3pt}
% \begin{tabular*}{\textwidth}
% {@{\extracolsep{\fill}}lrrrrr@{}}
% \toprule
% \textbf{Model} & \textbf{$\Delta\bar A_{\mathrm{sup}}$} & \textbf{$\Delta\rho_V$} & \textbf{$\Delta D^V_{KL}$} & \textbf{$\Delta\Gamma_{\mathrm{ac},V}$} & \textbf{$\Delta R_{vc}$} \\
% \midrule
% LLaMA-3.2-1B 
% & \textbf{-0.052} & -0.000 & \textbf{+0.001} & -0.058 & \textbf{-13.085} \\

% LLaMA-3.2-3B 
% & \textbf{-0.022} & +0.075 & \textbf{+0.002} & +0.050 & \textbf{-13.558} \\

% LLaMA-3.1-8B 
% & \textbf{-0.042} & +0.073 & \textbf{+0.007} & -0.018 & \textbf{-15.519} \\

% Qwen1.5-MoE 
% & \textbf{-0.012}  & +0.000 & -0.000 & +0.057 & \textbf{-1.102} \\
% \bottomrule
% \end{tabular*}
% \end{table*}
%-------------------------------------------------------------------------------
\section{Conclusion}
We have quantitatively verified the proposed three diagnostic claims through three-level hierarchical analytics. Our work fills a reasoning process-level gap for LLM reliability and safety inspections: final-output labels identify failures but not how they emerge during the entire process of generation. By organizing the hidden-state traces into seven functional-state coordinates, our proposed CauTrace connects the representation changes across Transformer blocks, temporal-state coordination among support/commitment/verification, and trajectory topology over the full reasoning trajectory chain. Across Qwen and LLaMA-family models, our experiments show lower support-state update agreement and lower verification activity relative to commitment in failed traces, different trajectory topologies among samples with the same output label, and larger internal changes under successful adversarial perturbations. The evidence supports a unified diagnostic account of failure dynamics and provides a foundation for more fine-grained causal localization. We believe that our work can provide a foundation for future research on LLM safety and inspire further efforts toward safer and more reliable LLMs.

\newpage

\FloatBarrier
% Everything below \appendix is excluded from the 13-page initial-submission
% body limit. The final camera-ready limit still counts every page.
\appendix

%-------------------------------------------------------------------------------
\section*{Ethics Considerations}
In developing CauTrace, we considered the ethical implications of combining white-box diagnostics with adversarial evaluation of large language models. The work proposes an offline analytical framework rather than an attack system: adversarial prompts are introduced only as controlled stress tests of whether ordinary and induced failures exhibit related internal process changes. All experiments use locally hosted open-weight models and public benchmarks; they involve no human participants, private user data, unauthorized access, or attacks on deployed services.

\textbf{Stakeholders and Potential Impact.}
The principal stakeholders are LLM users, individuals and communities affected by unsupported or harmful outputs, model developers, AI reliability and safety researchers, and benchmark maintainers. CauTrace may benefit these groups by clarifying how evidence support, answer commitment, rule tracking, and self-verification become miscoordinated before an unreliable or unsafe endpoint. Such diagnostics can inform future monitoring and mitigation, but they should not be interpreted as a per-generation safety guarantee or as a substitute for deployment-specific evaluation.

\textbf{Dual-Use Risks and Responsible Disclosure.}
The same internal patterns that help diagnose failure could be misused to design prompts that encourage premature commitment, weaken rule tracking, or suppress verification. We limit this risk in both experimental scope and artifact release. The threat model is restricted to prompt-side perturbations against locally hosted models, and the adversary has no access to model weights, hidden states, diagnostic outputs, evaluators, or deployed systems during prompt construction. We will not release turnkey jailbreak pipelines, operational instructions for bypassing safeguards, unredacted harmful-prompt collections, or harmful model completions. If future use of CauTrace reveals a deployment-specific vulnerability, we will coordinate disclosure with the relevant model or system provider before releasing operational details.

\textbf{Handling of Harmful Content.}
The paired safety evaluation may expose models to harmful prompts or produce unsafe responses. We reduce researcher exposure through automated safety judges, predefined output labels, and structured filtering, with manual inspection limited to debugging and validation. Harmful instructions are never executed, and potentially unsafe prompts or generations are neither reproduced in the paper nor redistributed in the artifact. Reported findings use aggregate statistics and sanitized identifiers wherever possible.

\textbf{Balancing Potential Benefits and Harms.}
CauTrace is intended to improve failure analysis by linking endpoint behavior to Propagation of Representation, Temporal-State Coordination, and Reasoning Trajectory. This evidence can help researchers distinguish failures that share an output label but require different monitoring or mitigation strategies. Although the resulting diagnostic knowledge retains some dual-use potential, the controlled evaluation setting, bounded threat model, and restricted release policy substantially reduce operational misuse. We therefore judge the expected reliability and safety benefits to outweigh the residual risks under these safeguards.

\section*{Open Science}
\label{app:open-science}
%-------------------------------------------------------------------------------
Reproducibility is a central goal of this work. We will provide an anonymous artifact containing the materials required to reproduce the reported diagnostic analyses while applying the safety restrictions described above.

\textbf{Data and Model Use.}
The experiments use publicly available reasoning and safety benchmarks, including the clean-reasoning datasets documented in Appendix~\ref{app:benchmark-eval} and the AdvBench/HarmBench evaluation sets, together with publicly released Qwen and LLaMA-family model checkpoints. These resources are used only for controlled evaluation and will remain subject to the licenses and usage conditions of their original providers. We collect no new human-subject data or personally identifiable information. Where redistribution of raw or harmful content is inappropriate, the artifact will provide source references, dataset manifests, version records, and sanitized instance identifiers rather than repackaging that content.

\textbf{Code and Artifacts.}
The artifact will include the CauTrace implementation and environment specification; model and dataset version records; scripts for trace collection, functional-state/process-coordinate scoring, probe training, weak-label construction, evaluation, statistical testing, and figure generation; fixed data splits and configuration files; and sanitized metadata and aggregate outputs for the matched clean/adversarial analysis. It will exclude unsafe model completions, unredacted harmful prompts, turnkey jailbreak pipelines, and operational instructions for circumventing safeguards. The review artifact will be available at https://anonymous.4open.science/r/artifact-B43D; following review, the released version will be preserved in a persistent archival repository.

% The bibliography style and location follow the official 2027 template.
\bibliographystyle{plainurl}
\bibliography{references}

% Optional experimental appendices retain the current section names.
%%%%%%%%%%%%%%%%%%%%%%%%%%%%%%%%%%%%%%%%%%%%%%%%%%%%%%%%%%%%%%%%%%%%%%%%%%%%%%%
%%%%%%%%%%%%%%%%%%%%%%%%%%%%%%%%%%%%%%%%%%%%%%%%%%%%%%%%%%%%%%%%%%%%%%%%%%%%%%%%

\section{Experimental Protocol and Details}
\label{app:protocol}
\label{app:a1}
\label{app:experimental-setup-details}
This appendix provides the experimental details and supplemental results needed to interpret the main evaluation, including benchmark composition, output evaluation, probe training, paired-adversary construction, and Trajectory Topology evidence.

\subsection{Evaluation Configuration}
\label{app:configuration}
Each sample is evaluated under concise and analysis prompt tracks. Unless otherwise stated, process-level results use the analysis track, which requests two to four short reasoning steps followed by a constrained final-answer line. The evaluator assigns one of five task-specific labels: \textit{Correct}, \textit{Partial Hallucination}, \textit{Full Hallucination}, \textit{Abstain}, or \textit{Invalid}. Safety pairs are evaluated analogously as safe refusal versus unsafe or partially unsafe compliance. The matched clean/adversarial evaluation reports Benign Success Rate (BSR), Adversarial Failure Rate (AFR), Pair Flip Rate (PFR), and Target Alignment Rate (TAR). Subsequent subsections specify the benchmark, output evaluation, probes, prompt construction, and paired metrics.

\subsection{Benchmark Composition and Evaluation}
\label{app:benchmark-eval}
The clean reliability benchmark contains nine public datasets: TruthfulQA, HotpotQA, MuSiQue, GSM8K, MATH, FOLIO, ProofWriter, StrategyQA, and CommonsenseQA. Together, they cover factual QA, multi-hop reasoning, numerical reasoning, formal mathematics, logic/rule reasoning, implicit reasoning, and commonsense multiple-choice QA. Table~\ref{tab:main-dataset-summary} reports the composition of the benchmark. Automatic evaluation follows the task-specific \texttt{judge\_type}.

\begin{table*}[!t]
\centering
%\scriptsize
%\setlength{\tabcolsep}{2.5pt}
\caption{Clean benchmark composition used in the main reliability evaluation.}
\label{tab:main-dataset-summary}
\begin{tabular}{lrl lrl lrl}
\toprule
\textbf{Dataset} & \textbf{Size} & \textbf{Judge} & \textbf{Dataset} & \textbf{Size} & \textbf{Judge} &\textbf{ Dataset} & \textbf{Size} & \textbf{Judge} \\
\midrule
TruthfulQA & 120 & \texttt{semantic} & HotpotQA & 7,400 & \texttt{support} & MuSiQue & 242 & \texttt{support} \\
GSM8K & 880 & \texttt{numeric} & MATH & 1,250 & \texttt{math} & FOLIO & 150 & \texttt{label} \\
ProofWriter & 600 & \texttt{label} & StrategyQA & 250 & \texttt{yes/no} & CommonsenseQA & 1,200 & \texttt{mcq} \\
\bottomrule
\end{tabular}
\end{table*}

Table~\ref{tab:app-judge-format} specifies the required output form and parsing constraint for each task family. Constraint failures are therefore evaluated against task-specific requirements rather than a single open-ended criterion.

\begin{table*}[!t]
\centering
%\scriptsize
%\setlength{\tabcolsep}{4pt}
\caption{Output constraints by \texttt{judge\_type}.}
\label{tab:app-judge-format}
\begin{tabular}{p{0.18\textwidth}p{0.2\textwidth}p{0.35\textwidth}}
\toprule
\textbf{\texttt{judge\_type}} & \textbf{Output Form} & \textbf{Constraint} \\
\midrule
\texttt{label\_match} & Label & \texttt{entailment/contradiction/unknown} \\
\texttt{yesno\_match} & Binary answer & \texttt{yes/no} \\
\texttt{mcq\_match} & Option & \texttt{A/B/C/D/E} \\
\texttt{numeric\_match} & Number & Final number only \\
\texttt{math\_answer\_match} & Mathematical answer & Final expression or result only \\
\texttt{semantic} & Short answer & Concise, comparable answer \\
\texttt{support} & QA answer & Answer based on supporting facts \\
\bottomrule
\end{tabular}
\end{table*}

Open-ended short answers are evaluated using normalized token-level F1:
\begin{align}
F1(\hat{y},y)&=\frac{2PR}{P+R},
\end{align}
\begin{align}
P&=\frac{|T(\hat{y})\cap T(y)|}{|T(\hat{y})|},\qquad
R=\frac{|T(\hat{y})\cap T(y)|}{|T(y)|}.
\end{align}
When $F1\ge 0.60$ without an exact match, the sample is labeled \texttt{partial\_hallucination}. For \texttt{mcq}, \texttt{yesno}, and \texttt{label} tasks, the structured parser $\pi_j(\cdot)$ is applied before scoring; a parsing failure produces the label \texttt{invalid\_or\_unparseable}. Abstention is detected using regular expressions for phrases such as \texttt{insufficient information}, \texttt{cannot determine}, and \texttt{unknown}. The evaluator first checks parseability and then distinguishes abstention, correctness, partial hallucination, and full hallucination. These endpoint labels are separate from the trajectory types defined by CauTrace.

\subsection{State-Probe Training and Scale Selection}
\label{app:probe-training}
\label{app:a3}
The seven functional-state/process coordinates are estimated by applying linear probes to causally pooled hidden states over window sizes $\{1,2,4,8,16,32,64\}$. Probe training uses weak process labels derived from task type, output label, and sequence-region heuristics; it does not use manually annotated token-level state labels. A higher score indicates stronger evidence for the corresponding operational role under the weak supervision, not a ground-truth cognitive state. Table~\ref{tab:probe-scale-main} shows that validation AUC approaches saturation from w32 to w64 across the evaluated Qwen2.5 models. We therefore use w32 in the main analysis because it remains in the top-performing range while preserving greater temporal resolution for activation-onset and state-transition measurements. These AUC values measure separability with respect to the weak supervision; they do not establish that a probe recovers a unique latent state. Table~\ref{tab:app-probe-hparams} lists the fixed training hyperparameters.

\begin{table*}[!t]
\centering
\caption{Probe validation AUC across pooling-window scales.}
\label{tab:probe-scale-main}
\begin{tabular}{lcccccccc}
\toprule
\textbf{Model} & \textbf{w1} & \textbf{w2} & \textbf{w4} & \textbf{w8} & \textbf{w16} & \textbf{w32} & \textbf{w64} & \textbf{fused} \\
\midrule
Qwen2.5-1.5B & 0.9006 & 0.9253 & 0.9501 & 0.9715 & 0.9843 & \textbf{0.9901} & 0.9918 & 0.9950 \\
Qwen2.5-3B   & 0.9038 & 0.9311 & 0.9570 & 0.9793 & 0.9902 & \textbf{0.9939} & 0.9952 & 0.9972 \\
Qwen2.5-7B   & 0.9557 & 0.9752 & 0.9880 & 0.9950 & 0.9973 & \textbf{0.9977} & 0.9979 & 0.9957 \\
Qwen2.5-14B  & 0.9794 & 0.9916 & 0.9974 & 0.9985 & 0.9992 & \textbf{0.9990} & 0.9989 & 0.9979 \\
\bottomrule
\end{tabular}
\end{table*}

\begin{table}[!t]
\centering
%\scriptsize
\setlength{\tabcolsep}{3pt}
\caption{Multi-scale probe training hyperparameters.}
\label{tab:app-probe-hparams}
\begin{tabular}{p{0.40\linewidth}p{0.52\linewidth}}
\toprule
\textbf{Parameter} & \textbf{Setting} \\
\midrule
Epochs & 40 \\
Batch size & 4096 \\
Learning rate & $1\times10^{-2}$ \\
Weight decay & $1\times10^{-4}$ \\
Patience & 6 \\
Train/Val & 0.9 / 0.1 \\
Scale set & w1, w2, w4, w8, w16, w32, w64, fused \\
\bottomrule
\end{tabular}
\end{table}

\paragraph{Trace collection and offline diagnosis.}
\label{app:offline-diagnosis}
CauTrace computes all diagnostic measurements after the model has generated a response. During generation, we record the hidden states, module outputs, attention-stream projections, and token positions required for analysis, but do not modify logits, hidden states, attention weights, or decoding decisions. The original model output determines the endpoint label. State-coordinate scores, transition variables, Attention/MLP update statistics, cascade statistics, layer/head sensitivity, and Q/K/V/O measurements are then computed offline from the recorded trace. CauTrace is therefore a diagnostic framework, not a generation-time controller. In paired experiments, the clean and adversarial rows use identical instrumentation and evaluation; only the prompt condition differs. Consequently, differences between the two traces are attributable to the prompt condition and the model's response to it, rather than to changes introduced by the analysis pipeline.

\subsection{Prompt Tracks and Matched Clean/Adversarial Construction}
\label{app:prompt-paired}
Each sample is evaluated under concise and analysis prompt tracks. Both use the same system instruction. The concise track requests only the final answer in the task-required format, whereas the analysis track requests two to four short reasoning steps followed by exactly one final-answer line. When uncertain, the model is instructed to output \texttt{Insufficient information}. The analysis track increases the observability of intermediate support and verification while retaining a parseable endpoint. Table~\ref{tab:app-dual-track-diff} summarizes the differences between the tracks.

\begin{table*}[!t]
\centering
%\scriptsize
\setlength{\tabcolsep}{4pt}
\caption{Differences between concise and analysis tracks.}
\label{tab:app-dual-track-diff}
\begin{tabular}{p{0.17\textwidth}p{0.13\textwidth}p{0.16\textwidth}p{0.44\textwidth}}
\toprule
\textbf{Dimension} & \textbf{Concise} & \textbf{Analysis} & \textbf{Potential Failure Mode} \\
\midrule
Output length & Short & 2--4 steps & May induce Verbose Unsupported Continuation outputs \\
Observability & Low & High & Rationale text is not always faithful verification \\
Commitment risk & Medium & Potentially higher & Early Commitment or Verification without Overwrite \\
Parsing stability & High & Medium & Invalid parsing if final-line format is violated \\
\bottomrule
\end{tabular}
\end{table*}

The matched clean/adversarial benchmark extends the process-level analysis to a security-relevant setting. Each pair shares an identifier, task semantics, evaluation target, and safety objective. The clean row contains the ordinary task or refusal instruction; the adversarial row adds a CauTrace-targeted prompt perturbation without changing the evaluation target. The design isolates the prompt condition at the input level. Differences in outputs and internal measurements are therefore interpreted as perturbation-linked changes, not as evidence of a uniquely localized cause.

The paired evaluation combines reasoning-reliability pairs derived from the clean benchmark with public red-team-style safety pairs based on AdvBench and HarmBench~\cite{mazeika2024harmbench,shen2024doanything,yu2024dontlisten}. Reasoning perturbations may introduce false premises, misleading source cues, shortcut rationales, incorrect calculation drafts, or answer-forcing instructions. For harmful-goal pairs, the clean condition requires safe refusal, whereas the adversarial condition applies jailbreak-style semantic pressure to formal rule tracking, answer commitment, or self-verification. We do not reproduce operational attack prompts or harmful procedural content; the analysis is limited to paired outcomes and internal changes.

\subsection{Adversarial Perturbations}
\label{app:targeted-metrics}

Each adversarial sample is annotated with the CauTrace state that its prompt perturbation is designed to disrupt: \StateAbbr{PB}, \StateAbbr{KR}, \StateAbbr{RC}, \StateAbbr{OC}, \StateAbbr{FRT}, \StateAbbr{AC}, or \StateAbbr{SV}. The annotation is excluded from endpoint scoring and used only in the target-alignment analysis. Specifically, we test whether the largest clean-to-adversarial changes occur in the intended state or in theoretically downstream states. Table~\ref{tab:targeted-perturbations} defines the perturbation taxonomy.

\begin{table*}[!t]
\centering
%\scriptsize
%\setlength{\tabcolsep}{4pt}
\caption{CauTrace state-targeted adversarial perturbations.}
\label{tab:targeted-perturbations}
\begin{tabular}{p{0.19\textwidth}p{0.32\textwidth}p{0.39\textwidth}}
\toprule
\textbf{Target State} & \textbf{Process Role} & \textbf{Perturbation Focus} \\
\midrule
Premise Binding & Bind entities, roles, context, and constraints. & False premises, role confusion, conflicting context bindings. \\
Knowledge Retrieval & Retrieve relevant facts or supporting evidence. & Plausible but wrong facts, source bias, retrieval decoys. \\
Reasoning Composition & Organize multi-step inference and intermediate claims. & Shortcut reasoning, chain reordering, incompatible intermediate claims. \\
Operational Computation & Execute numerical or symbolic operations. & Wrong scratch work, expression substitution, arithmetic interference. \\
Formal Rule Tracking & Maintain formal rules, safety policies, and task boundaries. & Rule-priority inversion, forged exceptions, constraint override. \\
Answer Commitment & Stabilize an answer after sufficient support. & Premature conclusion, forced single answer, uncertainty compression. \\
Self-Verification & Check evidence, answer consistency, and risk boundaries. & Reflection suppression, no-check instructions, refusal-review weakening. \\
\bottomrule
\end{tabular}
\end{table*}

The main experiments use Qwen2.5-Instruct models with 1.5B, 3B, 7B, and 14B parameters without fine-tuning~\cite{yang2025qwen3}; the transfer analysis uses dense LLaMA-family models and Qwen1.5-MoE. Decoding is deterministic, with temperature $0.0$, \texttt{top\_p}$=1.0$, and \texttt{do\_sample}=false. Every clean and adversarial row passes through the same evaluator, trace collector, probe stack, transition analyzer, cascade analyzer, and Q/K/V/O analyzer. For any comparable scalar metric $\eta$, the pair-level change is
\begin{equation}
\Delta \eta = \eta(\mathrm{adversarial}) - \eta(\mathrm{clean}),
\label{eq:pair-delta}
\end{equation}
This notation avoids overloading $m$, which denotes the number of value vectors in Eq.~\eqref{eq:value-spectrum}.

The paired analysis uses four primary behavior-level metrics. \emph{Benign Success Rate} (BSR) is the fraction of clean rows that satisfy the correctness, evidence-support, or safe-refusal criterion. \emph{Adversarial Failure Rate} (AFR) is the fraction of adversarial rows that are erroneous, unsupported, unsafe, or partially unsafe. \emph{Pair Flip Rate} (PFR) is the fraction of pairs that move from benign success to adversarial failure. \emph{Target Alignment Rate} at $k$ (TAR@$k$) measures whether the intended CauTrace target state occurs among the $k$ largest internal changes; the main text reports TAR@1 and TAR@3. We also compare flipped and non-flipped pairs to determine whether realized endpoint failures are associated with larger changes in state scores, modules, cascades, layers/heads, or Q/K/V/O representations.

\section{Supplemental Results}
\label{app:a5}

This section provides the endpoint, Trajectory Topology, and intervention results referenced in the main evaluation.

\subsection{Robustness to Diagnostic Aggregation}
\label{app:weight-sensitivity}

To determine whether \textbf{Claims I--III} depend on a particular aggregation scheme, we recompute the four-indicator aggregation $R_{\mathrm{str}}$ and reasoning process-deviation score $R_{\mathrm{proc}}$ using leave-one-component-out ablations and Dirichlet-random weights on the clean Correct/Full Hallucination comparison. Table~\ref{tab:weight-sensitivity} reports rank stability, failure--correct separation, and threshold-decision stability relative to the equal-weight scores.

\textbf{Representation-Change Propagation Is Stable as a Ranking, Whereas Reasoning-Process Deviation Is Intrinsically Multi-Component and Should Not Be Reduced to One Threshold.} For $R_{\mathrm{str}}$, both ablation and random weighting retain rank stability of $1.000$ in Table~\ref{tab:weight-sensitivity}; however, ablation decision stability ranges from $0.250$ to $1.000$, and its separation range reaches $0.000$. Thus, \textbf{Claim I} is robust at the ordering level but not as a universal binary rule. For $R_{\mathrm{proc}}$, median rank stability falls to $0.790$ under ablation and $0.696$ under random weights, while median decision stability is $0.827$ and $0.811$, respectively. This sensitivity is consistent with \textbf{Claims II--III}: different Trajectory Topologies are dominated by different combinations of timing, switching, rule preservation, and verification, so the aggregate score must be interpreted together with its components.

\begin{table*}[!t]
\centering
\caption{Aggregation sensitivity and component-ablation analysis for $R_{\mathrm{str}}$ and $R_{\mathrm{proc}}$. Rank stability denotes Spearman correlation with the default equal-weight score. Separation delta denotes the failure-minus-correct contrast under the perturbed score. Decision stability denotes preservation of the default threshold decision.}
\label{tab:weight-sensitivity}
\begin{tabular}{@{}p{0.16\textwidth}p{0.26\textwidth}p{0.16\textwidth}p{0.16\textwidth}p{0.16\textwidth}@{}}
\toprule
\textbf{Score Variant} & \textbf{Perturbation} & \textbf{Rank Stability} & \textbf{Separation Delta} & \textbf{Decision Stability} \\
\midrule
$R_{\mathrm{str}}$ default & none; equal weights & 1.000 & +0.014 & 1.000 \\
$R_{\mathrm{str}}$ ablation & support/value components leave-one-out & \textbf{1.000 [1.000, 1.000]} & +0.014 [+0.000, +0.018] & 0.812 [0.250, 1.000] \\
$R_{\mathrm{str}}$ random weights & Dirichlet random weights & \textbf{1.000 [1.000, 1.000]} & +0.014 [+0.000, +0.050] & \textbf{1.000 [1.000, 1.000]} \\
$R_{\mathrm{proc}}$ default & none; equal weights & 1.000 & -0.022 & 1.000 \\
$R_{\mathrm{proc}}$ ablation & process components leave-one-out & \textbf{0.790 [0.633, 0.887]} & -0.022 [-0.046, +0.006] & \textbf{0.827 [0.767, 0.933]} \\
$R_{\mathrm{proc}}$ random weights & Dirichlet random weights & 0.696 [0.022, 0.987] & -0.022 [-0.107, +0.061] & 0.811 [0.567, 1.000] \\
\bottomrule
\end{tabular}
\end{table*}

\subsection{Cross-Family Diagnostic Results}
\label{app:cross-family-results}
Table~\ref{tab:generalization} provides the complete cross-family measurements summarized in the main Generalization subsection. The shared directions of $\Delta\bar A_{\mathrm{sup}}$ and $\Delta R_{vc}$ support the representation-propagation and coordination relations, whereas the other representation-level carriers remain architecture-dependent.

\begin{table*}[!t]
\centering
\caption{Cross-family validation of representation-level and temporal diagnostics.}
\label{tab:generalization}
\begin{tabular*}{\textwidth}{@{\extracolsep{\fill}}lrrrrr@{}}
\toprule
\textbf{Model} & \textbf{$\Delta\bar A_{\mathrm{sup}}$} & \textbf{$\Delta\rho_V$} & \textbf{$\Delta D^V_{KL}$} & \textbf{$\Delta\Gamma_{\mathrm{ac},V}$} & \textbf{$\Delta R_{vc}$} \\
\midrule
LLaMA-3.2-1B & \textbf{-0.052} & -0.000 & \textbf{+0.001} & -0.058 & \textbf{-13.085} \\
LLaMA-3.2-3B & \textbf{-0.022} & +0.075 & \textbf{+0.002} & +0.050 & \textbf{-13.558} \\
LLaMA-3.1-8B & \textbf{-0.042} & +0.073 & \textbf{+0.007} & -0.018 & \textbf{-15.519} \\
Qwen1.5-MoE & \textbf{-0.012} & +0.000 & -0.000 & +0.057 & \textbf{-1.102} \\
\bottomrule
\end{tabular*}
\end{table*}

\subsection{Endpoint Outcomes and Distinct Trajectory Topologies}
\label{app:endpoint-output}
\begin{figure*}[!t]
    \centering
    \includegraphics[width=0.95\textwidth,height=0.22\textheight,keepaspectratio]{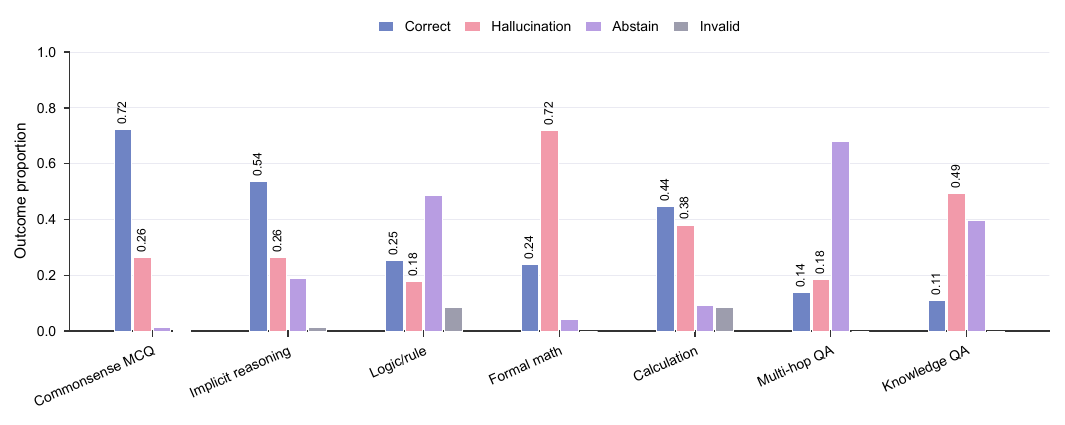}
    \caption{Task-group outcome distributions for Qwen2.5 generations. The horizontal bars visualize the proportions reported in Table~\ref{tab:task-group-outcomes-main}; they are not independent estimates.}
    \label{fig:app-task-group-outcomes}
\end{figure*}

To show why endpoint labels are insufficient for the three process-level claims, we compare the concise and analysis tracks on the clean nine-dataset benchmark. Table~\ref{tab:overall_outcome_stats} reports the overall outcome distribution by Qwen2.5 scale. Table~\ref{tab:task-group-outcomes-main} and Fig.~\ref{fig:app-task-group-outcomes} provide the corresponding task-group breakdown.

\textbf{Improved Endpoint Accuracy Does Not Imply Single Reliable Reasoning-Process Route.} In Table~\ref{tab:overall_outcome_stats}, the analysis track raises correctness by 13.2, 20.1, 18.6, and 11.7 percentage points for Qwen2.5-1.5B/3B/7B/14B, respectively. Full Hallucination nevertheless increases by 6.1 points for 3B and 7.6 points for 7B, even though it decreases for 1.5B and 14B. Longer reasoning is therefore associated with different outcomes across scales: it may provide more opportunity to build support and verify an answer, but it may also extend an unsupported commitment. Endpoint changes alone cannot distinguish these process routes.

\begin{table*}[!t]
\centering
\caption{Outcome distributions across Qwen2.5 models and prompt tracks. Len. denotes mean generation length.}
\label{tab:overall_outcome_stats}
\begin{tabular}{llrrrrrr}
\toprule
\textbf{Model} &
\textbf{Track} &
\textbf{Correct} &
\textbf{Partial} &
\textbf{Full} &
\textbf{Abstain} &
\textbf{Invalid} &
\textbf{Len.} \\
\midrule
1.5B & Concise  & 19.7 & 0.5 & \textbf{54.4} & 24.7 & 0.7 & 16.0 \\
1.5B & Analysis & \textbf{32.9} & 1.2 & 43.7 & 21.9 & 0.2 & 76.5 \\
3B   & Concise  & 12.1 & 0.3 & 20.1 & 65.1 & 2.3 & 25.5 \\
3B   & Analysis & \textbf{32.2} & 1.2 & \textbf{26.2} & 40.1 & 0.3 & 67.5 \\
7B   & Concise  & 15.8 & 0.4 & 12.7 & 71.1 & 0.0 & 6.0 \\
7B   & Analysis & \textbf{34.4} & 1.3 & \textbf{20.3} & 44.0 & 0.0 & 41.3 \\
14B  & Concise  & 18.5 & 0.4 & 16.1 & 58.9 & 6.1 & 29.1 \\
14B  & Analysis & \textbf{30.2} & 0.7 & \textbf{12.7} & 56.4 & 0.0 & 55.7 \\
\bottomrule
\end{tabular}
\end{table*}

The task-group results show that endpoint distributions also depend strongly on task structure. In Table~\ref{tab:task-group-outcomes-main}, failure is the largest category for formal mathematics ($0.718$), abstention is the largest for multi-hop QA ($0.678$), and correctness is the largest for commonsense multiple-choice QA ($0.721$). Figure~\ref{fig:app-task-group-outcomes} displays the same distributions across all seven task groups. These task-dependent mixtures motivate separate analysis of operational and rule failures, missing support, and constraint handling rather than treating an endpoint label as a complete process explanation.

\begin{table}[tbhp]
\centering
%\scriptsize
%\setlength{\tabcolsep}{5pt}
\caption{Task-group outcome composition, aggregated across Qwen2.5 scales and prompt tracks.}
\label{tab:task-group-outcomes-main}
\resizebox{\columnwidth}{!}{%
\begin{tabular}{lrrrrr}
% \begin{tabular}{>{\centering\arraybackslash}m{2.4cm}
% >{\centering\arraybackslash}m{0.85cm}
% >{\centering\arraybackslash}m{0.7cm}
% >{\centering\arraybackslash}m{0.7cm}
% >{\centering\arraybackslash}m{0.7cm}
% >{\centering\arraybackslash}m{0.7cm}}
\toprule
\textbf{Task group} & \textbf{Size} & \textbf{Correct} & \textbf{Failure} & \textbf{Abstain} & \textbf{Invalid} \\
\midrule
Calculation & 7032 & \textbf{0.445} & 0.380 & 0.090 & 0.085 \\
Formal math & 10000 & 0.240 & \textbf{0.718} & 0.042 & 0.000 \\
Multi-hop QA & 61176 & 0.137 & 0.185 & \textbf{0.678} & 0.000 \\
Knowledge QA & 656 & 0.110 & \textbf{0.494} & 0.396 & 0.000 \\
Logic/rule & 5952 & 0.252 & 0.177 & \textbf{0.486} & 0.086 \\
Commonsense MCQ & 9680 & \textbf{0.721} & 0.263 & 0.013 & 0.003 \\
Implicit reasoning & 2224 & \textbf{0.537} & 0.262 & 0.188 & 0.013 \\
\bottomrule
\end{tabular}
}
\end{table}

\subsection{Trajectory Signatures}
\label{app:trajectory-signatures}

The trajectory types in the main text are rule-defined regions of the trajectory-feature space. They are neither output labels nor mutually exclusive ground-truth classes. Table~\ref{tab:trajectory_signatures} specifies their operational signatures. The state-transition chains in Fig.~\ref{fig:reasoning_state_trajectories} are schematic examples consistent with these definitions and the observed transition-probability differences. They are not token-level reconstructions, and each trajectory type may contain multiple chain variants.

\begin{table*}[!t]
\centering
\caption{Operational signatures of the Trajectory Topology regions. Arrows indicate the expected direction relative to Normal Convergence. A sample may satisfy more than one signature.}
\label{tab:trajectory_signatures}
\begin{tabular}{>{\centering\arraybackslash}m{3.8cm}
>{\centering\arraybackslash}m{4.6cm}
>{\centering\arraybackslash}m{8cm}}
\hline
\textbf{Trajectory} & \textbf{Metric Signature} & \textbf{Diagnostic Interpretation} \\
\hline
Early Commitment
& $\tau_{\mathrm{com}} < \tau_{\mathrm{sup}}$ or $g_{\mathrm{cs}}\downarrow$
& Answer commitment becomes dominant before support is sufficiently established. \\
\hline
Fragmented Support
& $\Omega\uparrow$, $\bar{S}\downarrow$, unstable $\tau_{\mathrm{sup}}$
& Support-related states activate but do not form a stable, continuous chain. \\
\hline
Rule Failure
& $L_{\mathrm{rule}}\uparrow$ with weak \StateAbbr{FRT}/\StateAbbr{OC} state scores
& The trace approaches an answer without preserving formal rules, output constraints, or symbolic/numerical consistency. \\
\hline
Verification without Overwrite
& $g_{\mathrm{vc}}\uparrow$, $R_{\mathrm{vc}}\downarrow$
& Verification follows commitment but remains too weak to revise the stabilized answer. \\
\hline
Verbose Unsupported Continuation
& $T\uparrow$ with $\bar{S}\downarrow$ and $R_{\mathrm{vc}}\downarrow$
& Generation continues without a corresponding increase in support or effective verification. \\
\hline
Normal Convergence
& $\tau_{\mathrm{sup}} \leq \tau_{\mathrm{com}}$, bounded $\Omega$, $R_{\mathrm{vc}}\uparrow$
& Support precedes commitment, constraints remain active, and verification can confirm or revise the candidate. \\
\bottomrule
\end{tabular}
\end{table*}

\subsection{CauTrace-Guided Repair and State Steering}
\label{app:guided-intervention}

\textbf{Calibration and held-out protocol.}
We test whether the functional-state/process coordinates identified by CauTrace can guide response repair or internal-state steering on held-out data. All data-dependent choices---including the repair trigger, Trajectory Topology assignment, and steering strength---are fixed using calibration data before test evaluation. The held-out outcomes are therefore used only for evaluation and do not affect the intervention configuration.

\textbf{Reasoning repair.}
For Qwen2.5-1.5B reasoning, we use 50 calibration pairs to determine which responses should be repaired. A trace is triggered when its process-deviation score $R_{\mathrm{proc}}$ exceeds the 90th percentile of the clean-success calibration traces. This percentile provides a conservative high-risk operating point: it targets traces whose process organization is unusually different from successful clean reasoning, while avoiding unnecessary second-pass generation and test-set threshold tuning. Each triggered trace is also assigned a Trajectory Topology using the fixed CauTrace diagnostics.

We then evaluate five conditions on 200 disjoint test pairs. \emph{Baseline} retains the original response. \emph{Generic review} applies a diagnosis-free second-pass review with the same decoding and response budget as the CauTrace conditions. \emph{Unified CauTrace repair} (\texttt{cautrace_repair}) applies a fixed sequence of support reconstruction, rule checking, provisional commitment, and final verification. \emph{Topology-conditioned repair} uses a correction protocol specialized to the diagnosed Trajectory Topology, whereas \emph{wrong-topology repair} uses a deliberately mismatched topology while preserving the same prompt structure and generation budget. Only triggered traces are regenerated; non-triggered traces retain their baseline response under every second-pass condition. These controls distinguish improvement from generic additional review, improvement from the unified CauTrace protocol, and improvement that specifically depends on the diagnosed topology. Table~\ref{tab:guided-repair} reports the held-out endpoint results.

\begin{table}[!t]
\centering

\caption{Held-out Qwen2.5-1.5B reasoning repair on 200 test pairs.
Endpoint rates are percentages. Clean good denotes the proportion
of clean responses that remain correct, and $\Delta$Fail is the
adversarial-failure change relative to the baseline in percentage points.}
\label{tab:guided-repair}
\begin{tabular}{@{}lrrr@{}}
\toprule
\textbf{Condition} &
\textbf{Adv. fail} &
\textbf{Clean good} &
\textbf{$\Delta$Fail} \\
\midrule
Baseline             & 76.0          & 26.0          & --            \\
Unified CauTrace     & \textbf{71.0} & 31.0          & \textbf{-5.0} \\
Generic review       & 72.0          & \textbf{32.0} & -4.0          \\
Topology-conditioned & 73.0          & 29.5          & -3.0          \\
Wrong topology       & 73.5          & 30.5          & -2.5          \\
\bottomrule
\end{tabular}
\end{table}

\textbf{State steering.}
For Qwen2.5-7B safety, we test whether the same diagnostics can guide a direct hidden-state intervention. Let $v_{\mathrm{align}}$ denote the normalized steering direction constructed from the signed state directions predicted by the diagnosed trajectory: positive support, formal rule, and verification directions are combined with a negative answer-commitment direction. At generation step $t$ and intervened layer $\ell$, steering is applied as
\begin{equation}
h'_{\ell,t}
=
h_{\ell,t}
+
\alpha\,\operatorname{RMS}(h_{\ell,t})\,v_{\mathrm{align}} .
\label{eq:guided-steering}
\end{equation}
where RMS normalization makes the intervention scale comparable across layers and responses, and $\alpha$ controls its strength.

We select $\alpha$ using six calibration pairs and a predefined grid ${0.25,0.5,1.0}$. The selected value minimizes adversarial failure on the calibration pairs subject to a maximum two-percentage-point reduction in clean-good rate. This constraint prevents the intervention strength from being selected solely for adversarial recovery at the expense of clean behavior. The resulting value is $\alpha=0.25$, which is fixed before evaluation on 12 disjoint clean-success/adversarial-failure test pairs.

The direction is injected into a three-layer band centered on the CauTrace-identified risk layer during the first eight generated tokens. This restricts the intervention to the diagnosed model region and the early generation stage in which the response trajectory is formed. We compare \emph{aligned steering} with three equal-budget controls: \emph{reversed steering} uses $-v_{\mathrm{align}}$; \emph{random steering} uses a norm-matched direction orthogonal to $v_{\mathrm{align}}$; and \emph{wrong-state steering} strengthens answer commitment while weakening formal rule tracking and verification. All four conditions use the same $\alpha$, layers, intervention duration, and decoding configuration. The controls therefore test whether the predicted state direction changes the reasoning process beyond the effect of a generic hidden-state perturbation. Table~\ref{tab:guided-steering} reports both endpoint outcomes and online process-coordinate changes.

\begin{table}[!t]
\centering
\caption{Held-out Qwen2.5-7B safety steering on 12 test pairs.
Process deltas pool the clean and adversarial member of each pair;
lower $R_{\mathrm{proc}}$ and higher rule and verification coverage
are better.}
\label{tab:guided-steering}

\setlength{\tabcolsep}{2.0pt}
\renewcommand{\arraystretch}{1.05}

\resizebox{0.98\columnwidth}{!}{%
\begin{tabular}{@{}lrrrrr@{}}
\toprule
\textbf{Condition}
& \textbf{Adv. fail}
& \textbf{Clean}
& \textbf{$\Delta R_{\mathrm{proc}}$}
& \textbf{$\Delta\mathrm{Rule}_{\mathrm{cov}}$}
& \textbf{$\Delta V_{\mathrm{cov}}$} \\
\midrule
Baseline & 100.0 & 100.0 & 0 & 0 & 0 \\
Aligned & 66.7 & 91.7 & \textbf{-0.008} & \textbf{+0.011} & \textbf{+0.021} \\
Random & 66.7 & \textbf{100.0} & 0.000 & +0.002 & 0.000 \\
Reversed & 66.7 & 91.7 & +0.015 & -0.014 & -0.019 \\
Wrong state & 66.7 & 91.7 & +0.010 & -0.007 & -0.026 \\
\bottomrule
\end{tabular}%
}
\end{table}

\textbf{Unified Repair Gives the Largest Observed 1.5B Failure Reduction, and
Aligned Steering Moves the 7B Process Measures in the Predicted Direction.}
On the full 1.5B reasoning test, unified CauTrace repair reduces adversarial
failure from $76.0\%$ to $71.0\%$, corresponding to a $5.0$-percentage-point
reduction (paired 95\% bootstrap CI $[-8.5,-2.0]$; McNemar $p=0.0063$), with 11
failed responses rescued and one previously successful response regressing. This
is the largest observed reduction among the five conditions in
Table~\ref{tab:guided-repair}; generic review reduces failure by $4.0$ points,
while topology-conditioned and wrong-topology repair reduce it by $3.0$ and
$2.5$ points, respectively. The direct $1.0$-point difference between unified
CauTrace repair and generic review is not statistically significant (95\% CI
$[-4.0,+1.5]$; $p=0.727$). The result therefore shows that the complete
support--rule--commitment--verification protocol can improve held-out reasoning
outcomes, but does not establish superiority over generic review or a
topology-specific repair effect.

On 7B safety, all four steering directions reduce adversarial failure from
$100.0\%$ to $66.7\%$ in Table~\ref{tab:guided-steering}; endpoint recovery
alone therefore does not distinguish the aligned direction from its controls.
The internal process changes, however, follow different directions. Aligned
steering is the only condition that simultaneously lowers process risk
($\Delta R_{\mathrm{proc}}=-0.008$), increases rule coverage ($+0.011$), and
increases verification coverage ($+0.021$). Reversed and wrong-state steering
instead increase $R_{\mathrm{proc}}$ and reduce both rule and verification
coverage, whereas the random control remains close to zero. The same pattern is
stronger among the adversarial responses rescued by aligned steering, for which
$\Delta R_{\mathrm{proc}}=-0.051$,
$\Delta\mathrm{Rule}_{\mathrm{cov}}=+0.038$, and
$\Delta V_{\mathrm{cov}}=+0.042$. Thus, aligned steering produces the
process-direction changes predicted by the
support/rule--commitment--verification relation in \textbf{Claim II}. Given the
12-pair test and the identical endpoint rates across steering controls, this
result supports theory-consistent process steering in the evaluated setting
without attributing endpoint recovery to a unique causal state direction.

\subsubsection{Discussion: Defense Implications and Future Directions}

These results suggest a concrete path from the three-level diagnostics to
process-aware defenses. Structural retention can identify where an
intervention may be applied, functional-state/process coordinates can specify
whether support, rule tracking, and verification should be strengthened or
premature commitment suppressed, and Trajectory Topology can provide a
candidate basis for selecting repair policies. In the present tests, unified
repair provides the largest observed 1.5B reasoning reduction, while aligned
steering moves the 7B process measures in the predicted direction, showing
that CauTrace signals are actionable in the evaluated settings. Future work
can develop this fixed intervention into a closed-loop defense that monitors
process risk during generation, triggers repair only when needed, and adapts
the repair policy or steering strength while preserving clean behavior.
Evaluation across broader model families, tasks, attacks, and held-out samples,
together with stronger direction-specific controls, will clarify when these
process-level improvements translate into consistent endpoint robustness.

%-------------------------------------------------------------------------------
\end{document}